\documentclass[12pt]{article}
\usepackage{amsmath}
\usepackage{graphicx}
\usepackage{graphics}
\usepackage{algorithm}
\usepackage{algpseudocode}
\usepackage{algorithmicx}
\usepackage{natbib}
\usepackage{url} 
\usepackage{amsfonts,amsthm,amsmath,amsthm}
\usepackage{hyperref,natbib}
\usepackage{setspace}
\usepackage{graphicx}
\usepackage{hyperref}
\usepackage{bm}
\usepackage{autobreak}
\usepackage[usenames,dvipsnames,svgnames,table]{xcolor}

\newcommand{\C}{\mathbf{C}}
\newcommand{\G}{\mathbf{G}}
\newcommand{\M}{\mathbf{M}}
\newcommand{\X}{\mathbf{X}_n}
\newcommand{\R}{\mathbb{R}}
\newcommand{\Xset}{\mathcal{X}}
\newcommand{\m}{\mathbf{m}}
\newcommand{\x}{\mathbf{x}}
\newcommand{\y}{\mathbf{y}}
\newcommand{\K}{\mathbf{K}}
\newcommand{\Wij}{\mathbf{W}_{i,j}}
\newcommand{\Wijn}{\mathbf{W}_{i,j}^{(n)}}
\newcommand{\Wijnew}{\mathbf{W}_{i,j}^{(n+1)}}
\newcommand*{\Esp}[2]{\mathbb{E}_{#1}\left[#2\right]}
\newcommand{\A}{\mathcal{A}}

\newcommand{\resp}[2]{{ #2}}

\newcommand{\An}{\mathbf{A}}
\newcommand*{\Cov}[2]{\mathbb{C}ov_{#1}\left[#2\right]}
\newcommand*{\Var}[2]{\mathbb{V}ar_{#1}\left[#2\right]}

\newcommand{\xnew}{\tilde{\x}}
\newcommand{\vecyn}{\mathbf{y}_n}
\newcommand{\vecw}{\mathbf{w}}
\newcommand{\vecg}{\mathbf{g}}
\newcommand{\veck}{\mathbf{k}} 
\newcommand{\veckn}{\mathbf{k}_n} 

\newcommand{\nvar}{m}
\newcommand{\CGPn}{\C^{(n)}}
\newcommand{\CGPnew}{\C^{(n+1)}}

\newtheorem{Theorem}{Theorem}
\newtheorem{Corollary}{Corollary}

\newcommand{\blind}{0}

\begin{document}
	
	\def\spacingset#1{\renewcommand{\baselinestretch}%
		{#1}\small\normalsize} \spacingset{1}

	
	\if0\blind
	{
		\title{\bf Sequential Learning of Active Subspaces}
		\author{Nathan Wycoff\thanks{
				Current institution: Department of Statistics, Virginia Tech, Blacksburg, VA. Corresponding author: 
				\href{mailto:nathw95@vt.edu}{\tt nathw95@vt.edu}} \thanks{Mathematics and Computer Science Division, Argonne National Laboratory, Lemont, IL}\
			\and
				Micka\"el Binois\thanks{Current institution: Universit\'e C\^ote d'Azur, Inria, CNRS, LJAD, France} \footnotemark[2]
			\and 
				Stefan M.\ Wild\footnotemark[2]}
		\maketitle
	} \fi
	
	\if1\blind
	{
		\bigskip
		\bigskip
		\bigskip
		\begin{center}
			{\LARGE\bf Title}
		\end{center}
		\medskip
	} \fi
	
	\bigskip
	\begin{abstract}
		In recent years, active subspace methods (ASMs) have become a popular
		means of performing subspace sensitivity analysis on black-box
		functions. Naively applied, however, ASMs require gradient evaluations
		of the target function. In the event of noisy, expensive, or
		stochastic simulators, evaluating gradients via finite differencing
		may be infeasible. In such cases, often a surrogate model is employed,
		on which finite differencing is performed. When the surrogate model is
		a Gaussian process, we show that the ASM estimator is available in
		closed form, rendering the finite-difference approximation
		unnecessary. We use our closed-form solution to develop acquisition
		functions focused on sequential learning tailored to sensitivity
		analysis on top of ASMs. We also show that the traditional ASM
		estimator may be viewed as a method of moments estimator for a certain
		class of Gaussian processes. We demonstrate how uncertainty on
		Gaussian process hyperparameters may be propagated to uncertainty on
		the sensitivity analysis, allowing model-based confidence intervals on
		the active subspace. Our methodological developments are illustrated
		on several examples.
		 
	\end{abstract}
	
	\noindent%
	{\it Keywords:}  Active learning, Gaussian processes, linear embedding,
	dimension reduction, uncertainty quantification
	\vfill
	
	\newpage
	\spacingset{1.5}
	
	\section{Introduction}
	
	Simulation-based computer experiments are pervasive in diverse fields
	ranging from engineering to social science. Some of these simulations are
	time--and resource--consuming to such an extent that carefully selecting
	the experiments to be run is a keystone concern. A standard technique in
	this context is to replace the expensive simulator by an inexpensive
	surrogate (e.g., a quadratic response surface, a radial basis function, or
	a Gaussian process (GP)). Such approaches have been shown to be efficient,
	with the downside that performance drops sharply when the number of
	variables determining the experiment becomes more than a few dozen.
	
	Many efforts deal with this ``curse of dimensionality" (as coined by
	\citep{Bellman2003}) from different points of view. \resp{1-3}{Global} sensitivity analysis
	is concerned with determining the relative importance of each variable
	with respect to a quantity of interest, using methods such as screening or
	measures of influence; we refer to the work of \citet{Iooss2015} for a
	review. Another option is to assume additional structure about the problem
	at hand, such as additivity. The idea is to decompose the multivariate
	function $f$ into a sum of functions each of fewer variables, decreasing
	the amount of data needed for reliable inference. For GPs, purely additive
	examples include those of \citet{Durrande2010} and \citet{Duvenaud2011},
	while combinations of variables are considered, for instance, by
	\citet{Durrande2013} and \citet{Sung2017}. Learning a latent space has
	also been attempted, for example with a generative topographic mapping
	\citep{Viswanath2011} or internally in the GP model \citep{Titsias2010}.
	Related to this latter method is detecting the presence of an
	\textit{active subspace} \citep{Constantine2015}, the framework in which
	this article operates (see Section~\ref{sec:asm} for a formal definition).
	The principle is to find the directions accounting for most of the
	variation in the quantity of interest and then to conduct a study with a
	surrogate on those main directions only. This approach has been shown to
	perform well on a variety of problems, is easy to interpret because it
	corresponds to learning important linear combinations of the inputs, and
	has a theoretical foundation. It additionally allows for some
	visualization when the number of directions appears (or is selected) to be
	small.	
	
	The methodology typically consists of a random sampling of the
	high-dimensional design space to identify the active subspace, followed by
	a parameter study therein. While some guidelines for the primary stage
	exist, the methodology remains unsatisfying for several reasons. First, it
	requires gradient observations, which are not always available for legacy
	simulators or noisy problems \citep{LMW2019AN}. Relying on
	finite-difference approximations is an option--even for noisy simulators
	\citep{More2012}--but it has been shown to be less desirable
	\citep{Palar2018}. Directional derivatives may be used as well, from a
	low-rank matrix recovery perspective
	\citep{Djolonga2013,Constantine2015b}, but still demanding more costly
	observations if not directly available. As a result, replacing these
	derivative observations with those from a surrogate (e.g., a GP) is a
	standard approach \citep{Fukumizu2014,Othmer2016,Palar2018}. One must,
	however, still quantify the additional amount of uncertainty introduced.
	Second, unless a surrogate is used, the learned subspace cannot benefit
	from experiments run during the second stage because of the assumption of
	independent and identically distributed (iid) gradient sampling (because,
	for instance, optimization will typically involve function evaluations
	clustered in specific locations).
	
	In the GP literature, several authors have bypassed the two-stage approach
	by directly estimating the directions of influence, treating those as
	additional hyperparameters; see, for example, the work of
	\citet{Garnett2014} and \citet{Tripathy2016}.
	As a byproduct, this approach does not require gradient observations
	(although such observations can be accommodated) and does not require iid
	samples. \resp{}{Also, \citet{Marcy2017} proposed a sophisticated 
	fully Bayesian estimation framework of the dimension reduction
	matrix (including its dimension), which entails a challenging
	 optimization over the Grasmannian or Stiefold manifold.} 
 	Nevertheless, the proposed estimation procedures
	become much more complicated, requiring a variety of approximations. A
	major benefit of not requiring iid samples for active subspace estimation
	is to enable sequential learning procedures, that is, adding optimally
	informative points with respect to a quantity of interest.

	In this article, we bridge the gap between two-stage and direct estimation
	procedures for GPs, alleviating some of the limitations of recent active
	subspace methods. Our contributions can be summarized as follows:
	\begin{itemize}
		\item We prove that the classical active subspace estimator is a
		method of moments estimator for a certain class of GP models, formally
		establishing a link where similarities have previously been noted.
		\item We provide a closed-form expression for the matrix defining the
		active subspace for a GP for popular covariance kernels, rendering the
		Monte Carlo (MC) finite differencing on top of a surrogate
		unnecessary;
		\item Our closed-form solution allows sequential design to be carried
		out efficiently, allowing us to gain better estimates of the active
		subspace with fewer black-box evaluations, a feature especially important 
		when these evaluations are computationally expensive;
		\item We show how interval estimates may be calculated quickly via MC
		on the active subspace by propagating uncertainty from the GP
		hyperparameters.
	\end{itemize} 
	
	This paper is organized as follows. Section~\ref{sec:background} presents
	the active subspace approach as well as background on GPs.
	Section~\ref{sec:methodology} contains our methodological contributions,
	before illustrating them empirically in Section~\ref{sec:experiments}. We
	present our conclusions in Section~\ref{sec:discussion}.

	\section{Background}
	\label{sec:background}
	
	We first review the active subspace and GP paradigms.
	
	\subsection{Active Subspaces}
	
	\label{sec:asm} 
	
    The concept of \textit{active subspaces} \citep{Constantine2014} is
    appealing for fields from uncertainty quantification (UQ) to inverse
    problems. Informally, active subspace methods (ASMs) involve determining
    along which directions a scalar function of many variables changes a lot
    on average, and along which directions it is almost constant on average.
    The ideal family functions with which to describe the ASMs is that of
    \textit{ridge functions}, (not to be confused with ridge regression, an
    unrelated $\ell_2$ regularization technique), which are defined as
    functions that vary \textit{uniquely} in certain directions. To be
    precise, fix $\x \in \Xset \subseteq \R^\nvar$ and $\mathbf{A} \in \R^{r
    \times \nvar}$. Then a ridge function $f:\R^\nvar \rightarrow \R$ is a
    function of the form
	\begin{equation}
	f(\x) = g(\mathbf{A} \x)
	\label{ridgefunc}
	\end{equation}	
    for some function $g:\R^r \rightarrow \R$. Typically, $r < \nvar$, so $g$
    acts on a lower dimension than $\nvar$. By discovering $\mathbf{A}$, one
    exposes the low-dimensional structure in $f$. These functions underlie
    projection pursuit methods; see, for instance, the work of
    \citet{Friedman1981}.

    Ridge functions are constant along $\mathbf{A}$'s kernel. ASMs provide a
    framework for operating under looser assumptions: instead of the function
    being constant along $\mathbf{A}$'s kernel, ASMs stipulate that the
    directional derivatives in directions belonging to this subspace are
    significantly smaller than those belonging to $\mathbf{A}$'s range and
    make this assumption in expectation rather than absolutely.
	
	Rigorous definition of the active subspace requires us to consider the
	matrix $\C$, defined as the expected outer product of the gradient:
	\begin{equation}
	\C = \int_{\Xset} (\nabla f) (\nabla f)^\top d\mu.
	\label{Cdef}
	\end{equation}
    Here, $\Xset\subseteq \R^\nvar$ is the domain (of interest) of $f$, and
    $\mu$ is an arbitrary measure. Little attention has been paid to
    specifying $\mu$ beyond the heuristic that the Lebesgue measure is in some
    sense the natural starting point for bounded $\Xset$, whereas the Gaussian
    is used for unbounded $\Xset$.
	
	In practice, $\C$ is estimated by using a simple Monte Carlo procedure:
	sample $\x_{i} \sim \mu$ for $i \in \{1, \ldots, M\}$, then calculate the
	gradient at each point $\x_i$ to obtain the estimator
	\begin{equation}
	\hat{\C} = \frac{1}{M}\sum_{i=1}^M (\nabla f(\x_i)) (\nabla f(\x_i))^\top.
	\label{Chatdef}
	\end{equation}
	The properties of this estimator, which we call the classical or MC
	estimator, have been analyzed by \citet{Constantine2015} and
	\citet{Holodnak2018}, along with guidelines for choosing $M$, the MC
	sample size.
	
    The matrix $\C$ contains information about the gradient's direction on
    average through its eigendecomposition. In certain cases, one observes a
    jump in the eigenvectors: a ``spectral gap." This is indicative of an
    active subspace, defined as the principal eigenspace corresponding to the
    eigenvalues prior to the jump.
	
    \resp{2-1}{Related to the ASM is the concept of Sufficient Dimension Reduction (SDR),
    which aims to find a subspace capturing the stochastic dependence between the independent variables
    and the response. 
    However, SDR is applied in the context of
    observational data analysis, whereas methods bearing the
    name active subspaces tend to have a UQ application in mind, where the experimenter chooses the
    inputs. Hence SDR has its roots in the multivariate statistics community.} 
	We briefly
    outline some of the most popular SDR methods and refer the reader to the
    work of \citet{yanyuan2013} for a more complete review.
	
	Sliced inverse regression \citep{li1991}, among the earliest methods,
	involves breaking the response into bins, taking the mean of each feature
	within each bin, and then performing principal component analysis on the
	bins to give important directions. \resp{}{\cite{Glaws2020} relate this technique to the ridge functions
	with which the ASM is concerned, and \cite{Li2016} discuss algorithms
	suitable to both situations when the function of interest is exactly low dimensional 
	as well as when it is only approximately so.}
	
    Arguing that eigenvectors corresponding to large eigenvalues represent
    important directions, \citet{li1992} presents a method for subspace
    dimension reduction achieved by estimating the expected Hessian $\nabla^2
    f$. Computation involves solving the generalized eigenproblem on matrices
    with size equal to the dimension of the input space. 
	
    Also typical have been methods based on estimating these or related
    quantities by using kernel smoothers. An important early article in this
    regard is that of \citet{samarov1993}, who considered estimation of
    several functions defined in terms of integrals, including the expected
    outer product of the gradient, defined exactly as $\C$, as well as the
    expected Hessian. In the same article, sample estimators are constructed
    for such quantities by using kernel-regression-based estimators of the
    gradient at each sample location. Practical use of these techniques is
    complicated by the need to choose a kernel bandwidth; and although methods
    exist to this end, none may be considered ``best," and the methods may
    lead to different results \citep{Ghosh2018}. \citet{Fukumizu2014}
    alleviate some of those shortcomings but still rely on a MC estimator of
    $\C$. Among kernel methods, GPs offer additional benefits by taking a more
    probabilistic point of view.

	\subsection{Gaussian Processes and Linear Embeddings}
	
    Gaussian Processes are a common Bayesian nonparametric technique; see, for
    example, \citet{Rasmussen2006} for an introduction. Given any set of pairs
    $(\x_1, y_1), \ldots, (\x_n, y_n)$ with $\x_i$ belonging to a set $\Xset$
    representing inputs and $y_i \in \R$ representing outputs of some
    real-valued function defined on $\Xset\subseteq \R^\nvar$, the GP models
    the output vector $\vecyn = \left(y_i \right)_{1 \leq i \leq n}$ as coming
    from a Gaussian distribution with a mean vector and covariance matrix
    dependent on $\X = (\x_i)_{1 \leq i \leq n}$. Typically, the mean is set
    to be a constant or a linear function of $\X$, and the covariance matrix
    is determined via a \textit{kernel function}, a positive definite function
    defined on $\Xset\times \Xset$, such that the covariance satisfies
    \resp{1-4}{$\Cov{\omega}{y_i, y_j} = k(\x_i, \x_j)$} \resp{}{(here $\omega$ is an element of the sample space $\Omega$)}. 
    GP regression (also known as kriging) may be interpreted as a Bayesian linear regression
    on an infinite-dimensional space determined by the chosen kernel function 
    \citep{Scholkopf2001}.
	
    For the purposes of this article, suppose we are given $n$ observations of
    a \resp{}{(possibly noisy)} black-box function 
\resp{}{mapping the input space $\Xset$ and the stochastic outcome space $\Omega$ to the reals} $f: \Xset \times \Omega \rightarrow \R$, possibly
    depending on an active subspace of dimension $r$. Based on this data \resp{}{$\A_n
    = \left(\x_i, y_i = f(\x_i, \omega_i)  \right)_{1 \leq i \leq n}$}, and considering a
    prior (zero-mean) Gaussian process $Y \sim GP(0, k(\cdot,\cdot))$,
    classical multivariate normal properties result in the GP predictive
    equations; that is, $Y|\A_n \sim \mathcal{N}(m_n(\cdot),
    k_n(\cdot,\cdot))$ with
	\begin{align}
	m_n(\x) &= \veck(\x) \K_n^{-1} \vecyn\\
	k_n(\x, \x') &= k(\x,\x') - \veck(\x) \K_n^{-1} \veck(\x')^\top,
	\label{eq:kn}
	\end{align}
	where $\veck(\x) = \left(k(\x, \x_i) \right)_{1 \leq i \leq n}$ and $\K_n
	= \left( k(\x_i, \x_j) \right)_{1 \leq i,j \leq n}.$

    Unfortunately, without modification, GPs are not generally suited for
    problems of more than a few tens of variables. A common remedy is to
    conduct some manner of dimension reduction and perform the kriging on this
    smaller space. Popular among these methods are linear embeddings, which
    involve projecting the input data onto a lower-dimensional subspace. This
    may be done randomly as by \citet{Wang2016} or may be treated as kernel
    hyperparameters. A rank-deficient matrix in GP kernels to reduce
    dimensionality is presented, e.g., by \citet{Vivarelli:1999}. That the
    matrix is rank-deficient means that its range has smaller dimension than
    the underlying space, resulting in dimension reduction.
    \citet{Tripathy2016} explicitly model the GP as existing in the
    low-dimensional space defined by $\mathbf{A}\x$ for some $\mathbf{A}$
    (though this is easily seen to be equivalent to putting $\mathbf{A}$ in
    the kernel) and propose a two-step approach by alternating between
    optimizing with respect to the matrix $\mathbf{A}$ and the GP kernel
    hyperparameters. Since $\mathbf{A}$ is constrained to be orthogonal for
    identifiability purposes, this involves a nontrivial optimization over
    Grassmann manifolds (see also \citep{Marcy2017}). \citet{Hokanson2018}
    describe a similar approach using a polynomial model on the active
    subspace for which the two-step algorithm simplifies. Also taking the
    hyperparameter route are \citet{Garnett2014}, who define an acquisition
    function for sequential design that minimizes uncertainty on the subspace.
    This work bears similarity to ours but differs substantially in that it
    fits a GP on a low-dimensional subspace to be estimated, whereas we fit a
    GP on the entire space and use properties of GPs to estimate the important
    subspace.
	
	\resp{2-2/3/4/5/6/7/8}{
	\subsection{Linking GPs and ASMs}
	
	GPs via linear embeddings (GPLEs) and ASMs have much in common. Although
	links between GPs and ASMs have been observed previously, here we
	establish the prior properties of the classical ASM estimator $\hat{\C}$ as a method of
	moments (MoM) estimator for the linear embedding matrix in GPLEs in
	Theorem \ref{th:mom} and its corollary.

    \begin{Theorem}
		Assume we are given a mean-zero Gaussian process with the stationary
		kernel
		$
		k(\x_i,\x_j) = \sigma^2 \exp\{-0.5 (\x_i-\x_j)^\top\mathbf{A}^\top\mathbf{A}(\x_i-\x_j)\}
		$,
		that is, an anisotropic Gaussian kernel with variance parameter
		$\sigma^2$ and Mahalanobis distance given by the matrix
		$\mathbf{A}^\top\mathbf{A}$, with $\mathbf{A} \in
		\R^{r\times \nvar}$.
		Let $f \sim \textrm{GP}(0, k(\cdot, \cdot))$. 
		Then,
		$
		\Esp{\omega}{(\nabla f(\x_i, \omega)) (\nabla f(\x_i, \omega))^\top} = \sigma^2 \mathbf{A}^\top\mathbf{A}
		$
		for any $\x_i$, where $\omega$ denotes the source of stochasticity in $f$.
		\label{th:mom}
    \end{Theorem}
	\begin{proof}
	
    Since the active subspace is defined on the gradients of the function of
    interest, it is useful to derive the GP implied by the above model on the
    gradient.
		
    That the stochastic process implied by a GP on a function's partial
    derivatives is also a GP is well known.
    In particular, the prior covariance implied on partial derivatives of the GP is given by second-order derivatives
    of the kernel function \citep[Chapter 9.4]{Rasmussen2006}:
	\begin{equation*}
		\Cov{\omega}{\frac{\partial f(\x,\omega)}{\partial x_{i,d}}, 
		\frac{\partial f(\x,\omega)}{\partial x_{j,e}}} =
		\frac{\partial^2 k(\x_{i}, \x_{j})}{\partial x_{i,d}\partial x_{j,e}}.
	\end{equation*}

	where $e,d\in\{1,\ldots,m\}$ give input variables.
	It is therefore useful to derive the Hessian of our kernel function. Given
	that the derivative is
	\begin{equation*}
		\frac{\partial k(\x_1, \x_2)}{\partial \x_1} = 
		- \sigma^2 \mathbf{A}^\top\mathbf{A}(\x_1 - \x_2)  
		\exp\left\{-0.5 (\x_1 - \x_2)^\top\mathbf{A}^\top\mathbf{A}(\x_1 - \x_2)\right\},
	\end{equation*}		
	the Hessian can be expressed as
	\begin{equation*}
			\frac{\partial^2 k(\x_1, \x_2)}{\partial \x_1\partial 
\x_2^\top} = 
		\sigma^2 \left(\mathbf{I} - \mathbf{A}^\top\mathbf{A}(\x_1 - 
\x_2)(\x_1 - \x_2)^\top\right)
		\mathbf{A}^\top\mathbf{A}
		\exp\left\{-0.5 (\x_1 - \x_2)^\top\mathbf{A}^\top\mathbf{A}(\x_1 - 
\x_2)\right\}.
	\end{equation*}
		
	Now turn to the quantity of interest, 
	\begin{equation*}
		\Esp{\omega}{\nabla f(\x,\omega)\nabla f(\x,,\omega)^\top} =
		\Var{\omega}{\nabla f(\x,\omega)} + \Esp{\omega}{\nabla f(\x,\omega)}\Esp{\omega}{\nabla f(\x,\omega)}^\top.
	\end{equation*}
	
	Using our zero-mean assumption for our GP, our gradient's mean is then
	determined to be zero as well
	\resp{2-6}{\footnote{$\Esp{\omega}{\frac{\partial f(\mathbf{x},\omega)}{\partial x_{d}}}=
			\frac{\partial \Esp{\omega}{f(\mathbf{x},\omega)}}{\partial x_{d}} =
			\frac{\partial 0}{\partial x_{d}} =
			0$}}.
	 This deletes the second term, leaving
	\begin{equation*}
		\Esp{\omega}{\nabla f(\x,\omega)\nabla f(\x,\omega)^\top} =
		\Var{\omega}{\nabla f(\x,\omega)}. 
	\end{equation*}
	
	which we can evaluate using the Hessian derived above:
	
	\begin{align*}
		\Var{\omega}{\nabla f(\x,\omega)} &= \Cov{\omega}{\nabla f(\x,\omega), \nabla f(\x,\omega)}  \\
		&= \sigma^2 (\mathbf{I} - \mathbf{A}^\top\mathbf{A}(\x - \x)(\x - \x)^\top)
		\mathbf{A}^\top\mathbf{A}
		\exp\{-0.5 (\x - \x)^\top\mathbf{A}^\top\mathbf{A}(\x - \x)\} \\
		&= \sigma^2 \mathbf{A}^\top\mathbf{A}.
	\end{align*}

	\end{proof}

	\begin{Corollary}
	For any sample $\{\x_i\}$, $\Esp{\omega}{\frac{1}{n} \sum_{i =1 }^{n} \nabla f(\x_i,\omega) \nabla f(\x_i,\omega)^\top} =  \sigma^2 \mathbf{A}^\top\mathbf{A}$.
	\end{Corollary}
	
	Note that no assumption was made on the measure with respect to which the inputs $\x$ were sampled. This is a consequence of our stationary assumption: we are effectively assuming that the directions of global importance are important at each input point. To do otherwise would require a nonstationary kernel. This section has covered estimation of GP hyperparameters via the active subspace estimator $\hat{\mathbf{C}}$. There have also been efforts to, somewhat conversely, estimate $\C$ from a standard Gaussian process without linear embeddings. We will review this in the next section and further develop it in the remainder of the article.

}
	\subsection{Surrogate-Assisted Active Subspace Estimation}
	
    Estimating the matrix $\C$ is not straightforward if gradients for the
    function $f$ are not available, as is often the case in practice. Along
    with finite-difference gradient estimation, several gradient-free methods
    have been proposed to deal with this issue given $n$ observations of
    design points stored in a matrix $\X$ and the function $f$ evaluated at
    those points stored in a vector $\vecyn$.
	
    Perhaps the simplest of these ideas is proposed by
    \citet{Constantine2015}: simply perform linear regression of $\vecyn$ upon
    $\X$ and look at the subspace generated by the estimated regression
    coefficient vector. This method, applicable only when a one-dimensional
    subspace is desired, was found to be adequate experimentally on several
    problems. Also proposed by \citet{Constantine2015} is the use of local
    linear models, together with methods for aggregating their coefficient
    vectors. \resp{2-9}{ Further, \cite{constantine2015sketch} show how to estimate the active subspace 
   	when the full gradient is unavailable but certain directional derivatives are known.}
	
    Another approach is to retain the MC estimator but to use surrogate
    modeling to estimate the gradient at that point. \citet{palar2017},
    \citet{Othmer2016} and \citet{Namura2017} use finite differencing on a GP
    or polynomial chaos expansion in order to estimate gradients at random
    points throughout the input space, using the MC estimate of $\C$.
	
    We now show that the GP assumptions make such a process unnecessary, since
    a closed-form estimate is available.

	\section{Methodology}
	\label{sec:methodology}
	
	We now derive a closed-form estimator and a novel methodology for its use
	in sequential procedures. Our analytic estimator eases the computational
	burden and gives room for other procedures to be built on it.
	
	\subsection{Closed-Form Estimator of \texorpdfstring{$\C$}{C} from GPs}
	\label{sec:CGP}
    Recall our fundamental inference task of estimating $\C =  \Esp{\x}{(\nabla f)
    (\nabla f)^\top}$ given a set of observations $(\X, \vecyn)$. Although
    $\mu$ will be the Lebesgue measure in our experiments (Section
    \ref{sec:experiments}), we will neither make nor need the assumption that
    $\X \sim \mu$. Given our Gaussian assumptions on the stochastic process,
    the expectation is available analytically in terms of standard functions
    for popular kernels.

    Let us now express the matrix $\CGPn$ for a GP, starting \resp{2-10}{with 
    design points 
    $\A_n = \left(\x_i, y_i = f(\x)  \right)_{1 \leq i \leq n}$
	}
     before providing sequential expressions.
    Recall that $\K_n$ is the kernel matrix given $n$ observations: given a
    positive definite kernel function $k$, the components of $\K_n$ are given
    by $(\K_n)_{i,j} = k(\x_i, \x_j) + \tau^2 \delta_{i=j}$, where, in order
    to also account for noisy observations, $\delta$ is the Kronecker delta
    and $\tau^2$ is the error variance.
	
    Differentiation being a linear operator, it is well known that $\nabla Y =
    \partial Y / \partial x_1, \dots, \partial Y / \partial x_\nvar$ is also a
    Gaussian process. Assuming that k is twice differentiable, the joint
    distribution of $\left(Y(\X), \partial Y(\x) / \partial \x_1,
    \dots, \partial Y(\x) / \partial \x_\nvar \right)$ is 
	\[
	\begin{pmatrix}
	\vecyn\\
	\partial Y(\x) / \partial \x_1\\
	\vdots\\
	\partial Y(\x) / \partial \x_\nvar
	\end{pmatrix}
	\sim \mathcal{N}
	\left(
	\begin{pmatrix}
	\mathbf{0_n}\\
	0\\
	\vdots\\
	0
	\end{pmatrix},
	\begin{pmatrix}
	\K_n & \partial \veck(\x)^\top / \partial x_1 & \hdots & \partial \veck(\x)^\top / \partial x_\nvar\\
	\partial \veck(\x) / \partial x_1 & \partial^2k(\x, \x)/\partial x_1^2 & \hdots &  \partial^2k(\x, \x)/\partial x_1 \partial x_\nvar\\
	\vdots & \vdots & \ddots & \vdots\\
	\partial \veck(\x) \partial / x_\nvar & \partial^2k(\x, \x)/\partial x_\nvar \partial x_1 & \hdots & \partial^2 k(\x, \x)/\partial x_\nvar^2\\
	\end{pmatrix}
	\right),
	\]
	or, in shorthand,
	\[
	\begin{pmatrix}
	\vecyn\\
	\nabla Y (\x)
	\end{pmatrix}
	\sim \mathcal{N}
	\left(
	\begin{pmatrix}
	\mathbf{0_n}\\
	\mathbf{0_\nvar}
	\end{pmatrix},
	\begin{pmatrix}
	\K_n & \boldsymbol{\kappa}(\x)^\top\\
	\boldsymbol{\kappa}(\x) & \K_\nvar(\x,\x) 
	\end{pmatrix}
	\right),
	\]
	wherein $\boldsymbol{\kappa}_i(\x) = \frac{\partial \veck(\x)}{\partial x_i}\in\R^n$, for $i=1, \ldots, m$.


	As a result, $\nabla Y(\x)|\A_n \sim \mathcal{N}(\boldsymbol{\mu}_n(\x), \tilde{\K}_n(\x, \x))$ with 
	\begin{align*}
	\boldsymbol{\mu}_n(\x) &= \boldsymbol{\kappa}(\x) \K_n^{-1} \vecyn\\
	\tilde{\K}_n(\x, \x) & = \K_\nvar(\x, \x) - \boldsymbol{\kappa}(\x) \K_n^{-1} 
	\boldsymbol{\kappa}(\x)^\top,
	\end{align*} 
	which leads to a closed-form expression for $\C$, as given in Theorem \ref{th:C_GP}.
	\begin{Theorem}
        Let $k$ be a twice differentiable kernel, $\Wij :=
		\int_\Xset \boldsymbol{\kappa}_i(X)^\top \boldsymbol{\kappa}_j(X) d\mu$,
		and $E_{i,j} :=
		\int_\Xset \frac{\partial^2 k (X, X)}{\partial x_i \partial x_j}
		d\mu$. Then,  
		$C^{(n)}_{i,j} = E_{i,j} - tr\left( \K_n^{-1} \Wij \right) +
		\vecyn^\top \K_n^{-1} \Wij \K_n^{-1} \vecyn$.
		\label{th:C_GP}
	\end{Theorem}

	\begin{proof}
		Based on \citet[Equation (321)]{Petersen2008}, $\CGPn = \Esp{\x, \omega}{\nabla
		Y(X)(\nabla Y(X))^\top |\A_n } = \M + \m \m^\top$ with $X$ a
		random vector (e.g., same distribution as used for sampling the
		gradient),  $\m = \Esp{\x,\omega}{\nabla Y(X)|\A_n }$, $\M = \Var{\x,\omega}{\nabla
		Y(X)|\A_n }$. We use integration and expectations
		interchangeably.
		
		The first term is 
		$\m = \Esp{\x}{\boldsymbol{\mu}_n(X)} = \Esp{\x}{\boldsymbol{\kappa}(X)} \K_n^{-1} \vecyn 
		= \left[\int_{\Xset} \boldsymbol{\kappa}(\x) d\mu \right] \K_n^{-1} \vecyn$.
		For the second one, $\M$, we use the law of total covariance:
		\begin{align*}
		M_{i,j} =& \Cov{\x,\omega}{ \frac{\partial Y(X)}{\partial x_i}, 
		\frac{\partial Y(X)}{\partial x_j} |\A_n}\\
		=& \Esp{\x}{\Cov{\omega}{ \frac{\partial Y(X)}{\partial x_i}, 
		\frac{\partial Y(X)}{\partial x_j} |\A_n, X } } + \Cov{\x}{\Esp{\omega}{ 
		\frac{\partial Y(X)}{\partial x_i} | \A_n, X}, 
		\Esp{\omega}{\frac{\partial Y(X)}{\partial x_j} | \A_n, X} } \\ 
		=& \Esp{\x}{\tilde{K}_{i,j}^{(n)}(X,X)}+ \Cov{\x}{ \mu_{i}^{(n)}(X), \mu_{j}^{(n)}(X) }\\
		=&  \Esp{\x}{\frac{\partial^2 k (X, X)}{\partial x_i \partial x_j} - 
		\boldsymbol{\kappa}_i(X) \K_n^{-1} \boldsymbol{\kappa}_j(X)^\top} + 
		\Cov{\x}{\boldsymbol{\kappa}_i(X) \K_n^{-1} \vecyn,
		\boldsymbol{\kappa}_j(X) \K_n^{-1} \vecyn } \\
		=& E_{i,j} - \Esp{\x}{\boldsymbol{\kappa}_i(X) \K_n^{-1} \boldsymbol{\kappa}_j(X)^\top} 
		+ \vecyn^\top \K_n^{-1} \Cov{\x}{\boldsymbol{\kappa}_i(X),
		 \boldsymbol{\kappa}_j(X)}\K_n^{-1} \vecyn\\
		=& E_{i,j} - tr \left(\K_n^{-1} \left[\Cov{\x}{ \boldsymbol{\kappa}_i(X), 
		\boldsymbol{\kappa}_j(X) } + \Esp{\x}{\boldsymbol{\kappa}_i(X)}^\top 
		\Esp{\x}{\boldsymbol{\kappa}_j(X)} \right] \right) +\\
		& \vecyn^\top \K_n^{-1} \Cov{\x}{\boldsymbol{\kappa}_i(X), 
		\boldsymbol{\kappa}_j(X)} \K_n^{-1} \vecyn,
		\end{align*}
		where 
		$\boldsymbol{\kappa}_i(\x) = \Cov{\x}{ Y(\X), \frac{\partial Y (\x)}{\partial x_i}}$.
		
		The result follows by writing $C^{(n)}_{i,j} = M_{i,j} + m_i m_j$ and 
		denoting $\Wij = \Cov{\x}{ \boldsymbol{\kappa}_i(X), \boldsymbol{\kappa}_j(X)} + 
		\Esp{\x}{\boldsymbol{\kappa}_i(X)}^\top \Esp{\x}{\boldsymbol{\kappa}_j(X)} = \int_\Xset \boldsymbol{\kappa}_i(X) \boldsymbol{\kappa}_j(X)^\top d\mu$.
	\end{proof}
	
    Hence, once a form for the kernel is chosen \resp{2-12}{and the data have been observed,}
     the active subspace of the GP
    depends only on the kernel hyperparameters (e.g., the $\nvar$
    length scales), in contrast with the matrix parameterizations of
    \cite{Garnett2014} and \cite{Tripathy2016} that involve $\nvar \times r$
    hyperparameters. For classically used kernels such as tensor product
    versions of the Gaussian and Mat\'ern (with smoothness parameter $\nu \in
    \{3/2, 5/2\}$) kernels, $E_{i,j}$ and $\Wij$ are available even in closed
    form (see Appendix \ref{ap:kernelexps} and supplementary materials).
    $C^{(n)}_{i,j}$ can be decomposed into two parts: the first two terms
    correspond to an integrated mean square prediction error (especially
    apparent when $i = j$), while the second is related to the product of
    predictive means. \resp{2-12}{The matrix $\CGPn$ is thus estimated via 
    the (nonstationary, in contrast to Theorem \ref{th:mom}) posterior Gaussian process
    conditioned on the observed input-output pairs $\A_n$. }
    	
    
    The diagonal terms of the matrix can also be connected
    to derivative-based global sensitivity measures, for which the expressions
    have also been derived by \citet{DeLozzo2016}.

    Given our focus here on sequential design, also of interest is an update
    formula for calculating $\CGPnew$ (i.e., our estimate for $\C$ given
    $\A_{n+1}$) accounting for the fact that we have already calculated
    $\CGPn$ using $\K_n$ and $\Wijn$. Specifically, $\K_{n+1}$ and $\Wijnew$
    can be decomposed as
    \[
	\K_{n+1}=\begin{bmatrix}
	\K_{n} & \veckn(\xnew) \\
	\veckn(\xnew)^\top & k(\xnew,\xnew) + \tau^2
	\end{bmatrix}, \quad
	\Wijnew= \begin{bmatrix}
	\Wijn & \vecw_a(\xnew)\\
	\vecw_b(\xnew)^\top & w(\xnew, \xnew)
	\end{bmatrix}.
	\] 

    The resulting update formula is given in Theorem \ref{th:Cn1}.
	\begin{Theorem}
	Denote $\vecg(\xnew)= -\sigma_n^2(\xnew)^{-1}\K_n^{-1} \veckn(\xnew)$ and
	$\sigma^2_n(\xnew) = k_n(\xnew, \xnew)$. Given a new design point $\xnew$
	but not the function value at this location (i.e., $y_{n+1} \sim
	\mathcal{N}(m_n(\xnew), k_n(\xnew, \xnew))$), the random variable
	$C_{i,j}^{(n+1)} - C_{i,j}^{(n)}$  can be written as
	\begin{equation*}
	 C_{i,j}^{(n+1)} - C_{i,j}^{(n)} = \alpha_{i,j}(\xnew) + 
	 Z \beta_{i,j}(\xnew) + Z^2 \gamma_{i,j}(\xnew),
	\end{equation*}
	with $Z \sim \mathcal{N}(0,1)$ and
		\begin{align*}
	\alpha_{i,j}(\xnew) =& - (\vecw_a(\xnew) + \vecw_b(\xnew))^\top \vecg(\xnew) -
	 \frac{w(\xnew, \xnew) + 
	 \veckn(\xnew)^\top \K_n^{-1} \Wijn \K_n^{-1} \veckn(\xnew)}{\sigma^2_n(\xnew)}\\ 
	\beta_{i,j}(\xnew) =&  \frac{\vecyn^\top \K_n^{-1} 
	\Wijn \K_n^{-1} \veckn(\xnew) + \veckn(\xnew)^\top \K_n^{-1} \Wijn \K_n^{-1} 
	\vecyn - (\vecw_a(\xnew) + \vecw_b(\xnew))^\top \K_{n}^{-1} \vecyn}{\sigma_n(\xnew)} \\ 
	\gamma_{i,j}(\xnew) =&   \frac{w(\xnew, \xnew) 
	+ \veckn(\xnew)^\top \K_n^{-1} \Wijn \K_n^{-1} \veckn(\xnew)- (\vecw_a(\xnew) + 
	\vecw_b(\xnew))^\top \K_n^{-1} \veckn(\xnew)}{\sigma_n^2(\xnew)}.
	\end{align*}
	Once the response $\y_{n+1}$ has been observed, the expression evaluates to
	\begin{align*}
	&C_{i,j}^{(n+1)} - C_{i,j}^{(n)}= - (\vecw_a(\xnew) + \vecw_b(\xnew))^\top \vecg(\xnew)\\
	&- \left(\vecyn^\top \vecg(\xnew) + \frac{y_{n+1}}{\sigma_n^2(\xnew)}\right) \left[ \vecyn^\top \K_n^{-1} \Wijn \K_n^{-1} \veckn(\xnew) + \veckn(\xnew)^\top \K_n^{-1} \Wijn \K_n^{-1} \vecyn \right]\\ 
	&+ \left(\vecyn^\top \vecg(\xnew) + \frac{y_{n+1}}{\sigma_n^2(\xnew)}\right) (\vecw_a(\xnew) + \vecw_b(\xnew))^\top (\K_{n}^{-1} \vecyn + \vecg(\xnew) y_{n+1} - \vecg(\xnew) \veckn(\xnew)^\top \K_n^{-1} \vecyn)\\
	&+ \left[ \left(\vecyn^\top \vecg(\xnew) + \frac{y_{n+1}}{\sigma_n^2(\xnew)}\right)^2 - 
	\frac{1}{\sigma_n^2(\xnew)} \right] \left[ w(\xnew, \xnew) + \veckn(\xnew)^\top 
	\K_n^{-1} \Wijn \K_n^{-1} \veckn(\xnew) \right].
	\end{align*}
	\label{th:Cn1}
	\end{Theorem}
	\begin{proof}
	The derivation is detailed in Appendix \ref{sec:update}.
	\end{proof}

	Note that in contrast to the classical MC estimator $\hat{\C}$, by using a
	GP to predict everywhere, one can follow an ``off-policy" approach: design
	points $\X$ need not be sampled from $\mu$. Combined with the above update
	expression, this enables sequential design capabilities to learn active
	subspaces more efficiently.

	\subsection{Illustration: Comparison with ARD}
	\label{ard}
	
    At first glance, a disadvantage of kernel methods as compared with
    classical parametric models is the difficulty in interpreting model
    parameters. A popular heuristic for determining relative variable
    importance is that of automatic relevance determination (ARD)
    \citep[Chapter 5.1]{Rasmussen2006}, which asserts that input dimensions
    with large length scales are of little importance. This is due to the
    limiting behavior of many popular kernels that asymptotically ignore an
    input dimension as its length scale tends to infinity. This is taken
    advantage of, for example, for the purpose of axis-aligned dimension
    reduction for optimization \citep{Salem2018}. For two finite length scales
    $l_1 > l_2$ belonging to input variables $x_1, x_2$, however, there is no
    guarantee that $x_2$ is more important than $x_1$. As we illustrate here
    with a simple example, examining the loadings of variables in the active
    subspace eigenbasis may be more informative.
	
\begin{figure}[ht]
		\centering
		\includegraphics[width=\textwidth]{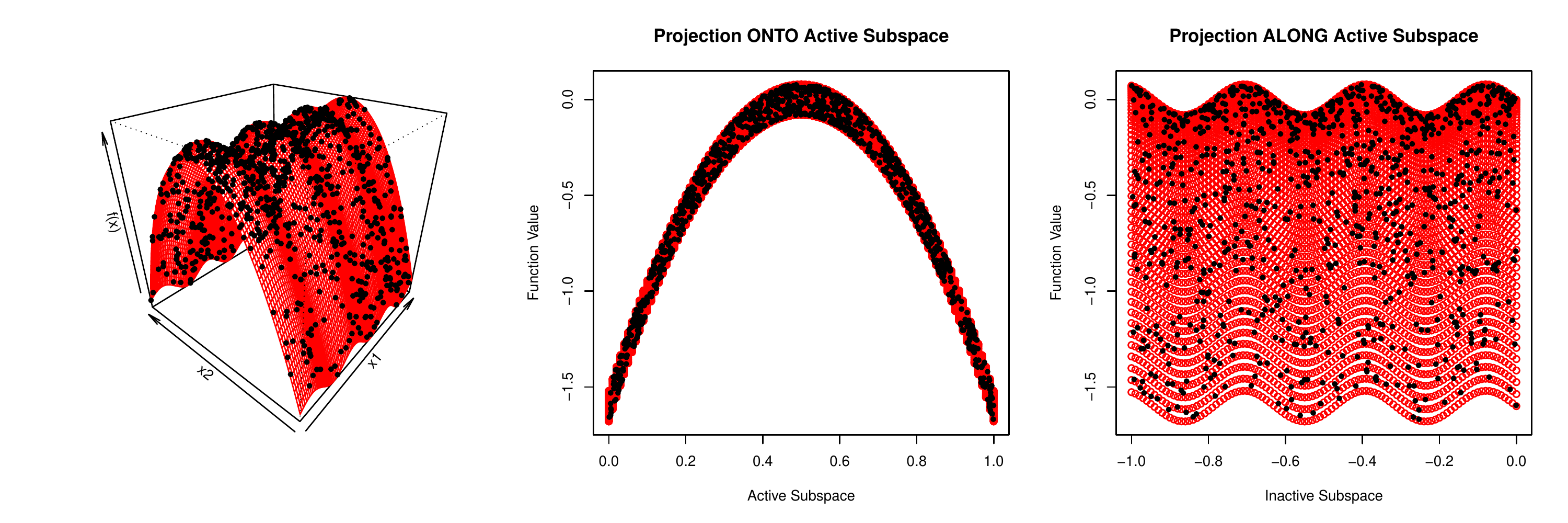}
		\caption{The 2-D function \resp{1-7}{given by Equation} \eqref{eq:testfun} 
		exhibits insignificant but high-frequency behavior in one dimension 
		and significant but long-range behavior in the other. Filled black
		points are design points while red circles represent points evaluated
		on a regular grid for visualization. The ASM
		reveals a linear subspace along which most of the important changes
		occur (center), while the ARD principle would select the
		high-frequency dimension (right).}
        \label{fig:ard}
	\end{figure}

    Consider the function 
    \begin{equation}
     f(x_1, x_2) = a \sin(bx_1) + c x_2^2; \qquad a = 0.1, \, b = 20, \, c = -4;
     \qquad (x_1,x_2) \in [0,1]^2
    \label{eq:testfun}
     \end{equation}
    illustrated in Figure~\ref{fig:ard}. Along the $x_1$ dimension is a
    sinusoidal function with high frequency but low amplitude, and along the
    other is simply a quadratic function in $x_2$, which dominates in terms of
    determining the function value. Given 1,000 observations uniformly
    distributed from the 2D unit interval, a GP with a Gaussian kernel gives
    length-scale estimates (determined via maximum likelihood estimation) of
    0.069 for $x_1$ and 0.37 for $x_2$, which, according to the ARD principle,
    mistakenly suggests $x_1$ is more important than $x_2$.
	
    Turning now to the active subspace estimated by the GP for the same
    sample, we find the eigendecomposition of our estimated $\C$ to be
	\begin{equation*}
		\hat{\C} = \begin{bmatrix}
		-0.00 & -0.99 \\
		0.99 & -0.00
		\end{bmatrix}
		\quad
		\begin{bmatrix}
		13.63 & 0 \\
		0 & 1.30
		\end{bmatrix}
		\quad
		\begin{bmatrix}
		-0.00 & 0.99 \\
		-0.99 & -0.00
		\end{bmatrix}.
	\end{equation*}
    The first eigenvector represents $x_2$ while the second represents $x_1$,
    and the first eigenvalue is an order of magnitude greater than the first,
    suggesting, correctly, that the variable $x_2$ is of greater importance
    \footnote{\resp{2-13}{A similar example was considered in \cite{constantine2017stationary}, where the magnitudes of $a$ and $c$ were switched, and the opposite conclusion of variable importance was reached.}}.
    The effect of projecting along the eigenvectors is further illustrated in
    Figure~\ref{fig:ard}. 
    \resp{1-6}{The ASM is not the only global sensitivity
    metric that chooses the ``right'' coordinate: indeed, the Sobol indices \citep{sobol2001} for our two variables are 
    approximately $0.003$ and $0.997$, respectively. 
    The power in the ASM lies in its ability to detect non-axis-aligned directions of importance.
    }
	
	\subsection{Quantifying the Uncertainty in C}
	
\begin{figure}[htpb]
    	\includegraphics[scale=0.45]{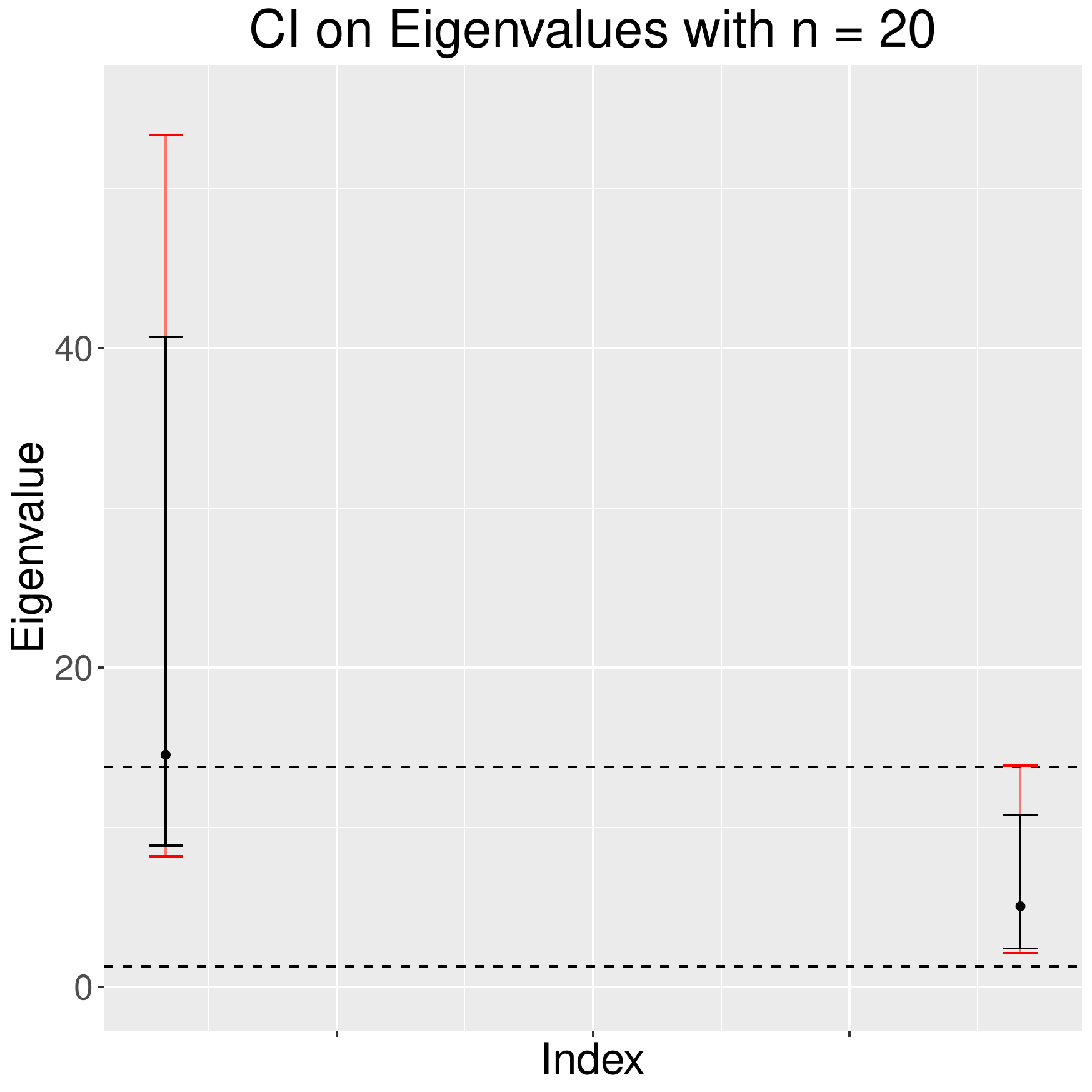}
    	\includegraphics[scale=0.45]{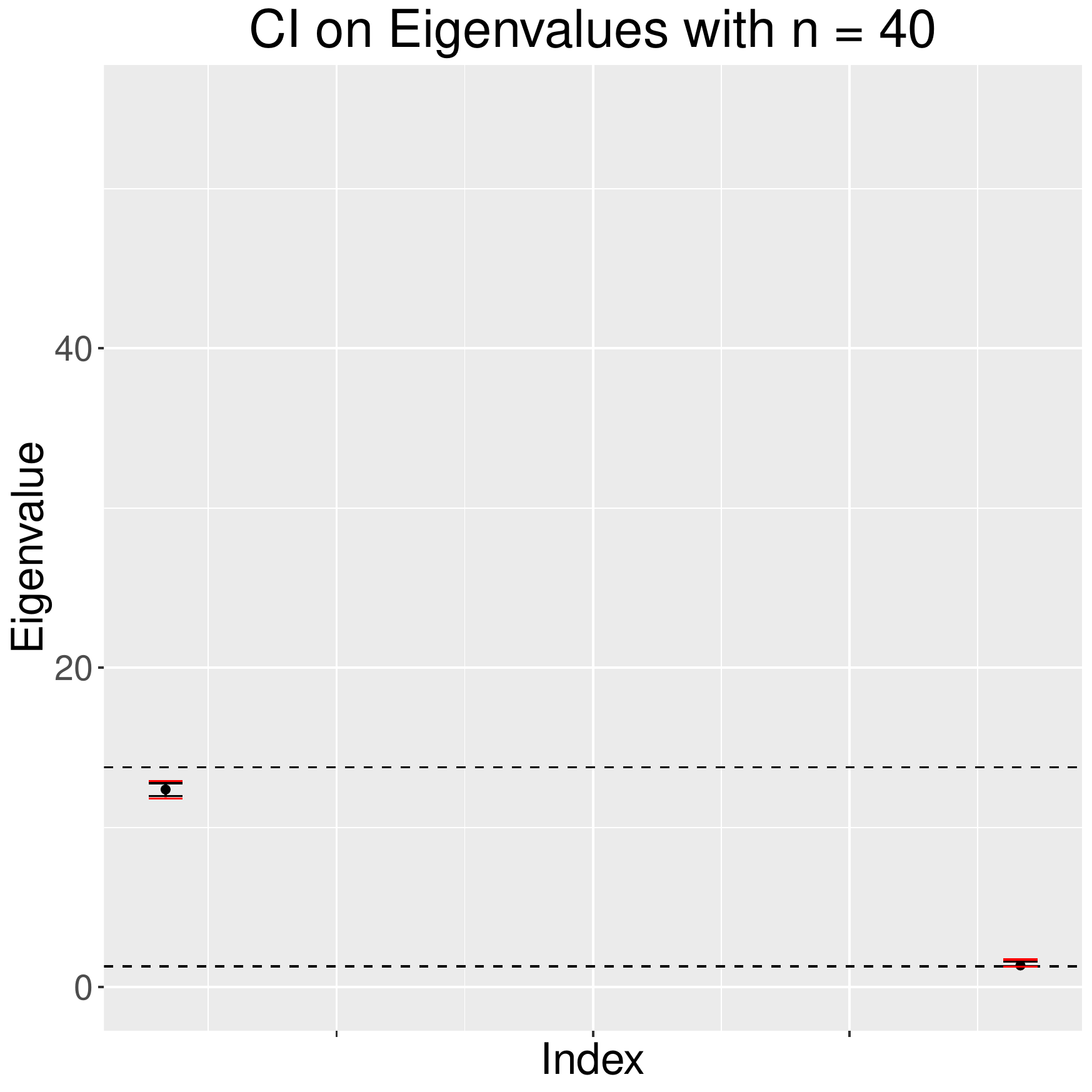}
	   	\caption{Eigenvalues from the problem in \eqref{eq:testfun}
	   	are given interval estimates, expressing decreasing model uncertainty
    	with samples of size 20 (left) and 40 (right). The Hessian expansion
    	of the log-likelihood at its maximum gives a correlation of 0.16
    	between the two length-scale parameters. 95 and 99\% confidence
    	intervals are given in black and red, respectively, while the
    	horizontal lines give the locations of the ``true" eigenvalues
    	estimated via classical MC techniques with 10,000 samples.}
	  	\label{fig:uq}
    \end{figure}

    Of perhaps equal importance as giving an estimate for $\C$ is quantifying
    uncertainty in that estimate. Given an estimate, \citet{Constantine2015}
    shows how the bootstrap method \citep{Efron1981} can be used to create
    intervals by applying parametric bootstrapping to simple methods used to
    estimate active subspaces (e.g., OLS, LL).
    However, application of the bootstrap to more sophisticated active
    subspace estimation methods such as the one presented in this article can
    lead to computational challenges. Instead, uncertainty of GP parameters
    can be propagated to uncertainty of $\C$ via a simple MC procedure. Myriad
    methods 
    for quantifying GP hyperparameter uncertainty exist: a full
    posterior can be developed by using MCMC procedures, approximate
    posteriors can be determined (e.g., via variational methods), or the
    asymptotic normality of maximum likelihood estimators can be exploited. In
    light of its computational ease, the last method is explored here. This
    method involves computing the Hessian of the log likelihood at its maximum
    value, giving a covariance matrix for the parameters, and treating the
    maximizing parameter setting as the mean. This uncertainty on model
    parameters may be propagated to the active-subspace-defining singular
    values via MC: for each drawn parameter setting, estimate the singular
    values, then keep, for instance, the middle 95 percent of the draws. As an
    illustration, in Figure~\ref{fig:uq} we form intervals on the eigenvalues
    of the problem defined in \eqref{eq:testfun}. The downside of this
    approach is that, while it is speedy, it does not account for uncertainty
    as well as would a fully Bayesian approach leveraging MCMC.
    
	\subsection{Acquisition Functions for Active Subspaces}
	
    Equipped with a closed-form GP estimate of $\C$ for a fixed design, we
    turn to the problem of choosing the next design to be evaluated given our
    observed responses so as to best learn about the active subspace. As do
    \citet{Labopin-Richard2016}, we take the approach of selecting the design
    that maximizes variance of our unknown quantity $\C$ at the next
    iteration. However, how to define this variance in our case is not clear.
    One could define subspace variance through ideas such as Fr\'echet
    variance \citep{frechet1948} and seek to minimize this, but the dimension
    of the active subspace is not known ahead of time, which would have to be
    determined by human observer or via a heuristic at each step of the
    sequential design process. Also possible is to consider measuring the
    variance of the spectral gap, which is problematic for the same reasons.
    We instead focus on the variance of the matrix $\CGPnew$, which we measure
    in three ways.
	
	Since examination of $\C$'s spectrum is central to making decisions
	regarding the dimension of the active subspace, pinning down $\C$'s
	eigenvalues is of great interest. As such, we define the \texttt{Trace}
	acquisition function using the variance of the mean eigenvalue, which may be
	optimized by maximizing the variance of the trace of $\CGPnew$:
	\begin{equation*}
		\texttt{Trace} = \Var{\resp{}{\omega}}{\textrm{tr}(\CGPnew)}.
	\end{equation*}
	
	Of practical concern is that the \texttt{Trace} function may be calculated
	without computing the off-diagonal elements of the $\C$ matrix, which
	seems to some degree antithetical to the principle of ASMs seeking
	non-axis-aligned directions of importance.
	
	
	We also consider two acquisition functions attempting to directly measure
	$\C$'s variance, which we term \texttt{Var1},
	\begin{equation*}
		\texttt{Var1} = \left\|\Esp{\resp{}{\omega}}{\left(\CGPnew - \Esp{\resp{}{\omega}}{\CGPnew}\right)\odot\left(\CGPnew - 
		\Esp{\resp{}{\omega}}{\CGPnew}\right)}\right\|_F^2,
	\end{equation*}
	 and \texttt{Var2},
	\begin{equation*}
		\texttt{Var2} = \left\|\Esp{\resp{}{\omega}}{\left(\CGPnew - \Esp{\resp{}{\omega}}{\CGPnew}\right)\left(\CGPnew - 
		\Esp{\resp{}{\omega}}{\CGPnew}\right)}\right\|_F^2.
	\end{equation*}
    Here, $\odot$ represents elementwise (Hadamard product) multiplication and
    $\|\mathbf{A}\|_{F}$ denotes the Frobenius norm of $\mathbf{A}$ (the root
    of the sum of its squared elements).

    Of practical interest is that, again, closed-form expressions can be
    obtained for these three acquisition functions.

	\begin{Theorem}
	Given the coefficients $\beta_{i,j}(\xnew)$ and $\gamma_{i,j}(\xnew)$ from
	Theorem~\ref{th:Cn1} stored in matrices $\mathbf{B}(\xnew)$ and
	$\bm{\Gamma}(\xnew)$, the acquisition functions and their gradients are
	available in the closed forms shown in Table~\ref{tab:acq_expr}.
	\label{th:anltAcq}
	\end{Theorem}
	\begin{proof}
		The derivation is detailed in Appendix~\ref{sec:infill}. 
	\end{proof}

	We note further that $\mathbf{B}(\xnew)$ and $\bm{\Gamma}(\xnew)$ have
	closed-form expressions for popular kernels, including the Gaussian,
	Mat\'ern 3/2, and Mat\'ern 5/2; see Appendix \ref{ap:kernelexps}.

	\begin{table}[h]
		\centering
		
		\caption{Analytic expressions and gradients with respect to the new
		location $\xnew$ for each of our acquisition functions. For
		compactness, we employ $\An=\CGPnew - \Esp{\resp{}{\omega}}{\CGPnew}$. Dependence of
		$\mathbf{B}(\xnew), \partial\mathbf{B}(\xnew), \bm{\Gamma}(\xnew)$,
		and $\partial\bm{\Gamma}(\xnew)$ on $\xnew$ has been suppressed for
		brevity. \label{tab:acq_expr}}
	
	\vspace{1em}

		\resizebox{\columnwidth}{!}{
			\begin{tabular}{l*{3}{c}}
				\textbf{Name}              & \textbf{Definition} & 
				\textbf{Expression} & \textbf{d-Derivative}\\
				\hline
				\texttt{Trace} & $\Var{\resp{}{\omega}}{\textrm{tr}(\C^{(n+1)})}$  & 
				$\textrm{tr}(\mathbf{B}\odot\mathbf{B}) + 
				2\textrm{tr}(\bm{\Gamma}\odot\bm{\Gamma})$ & 
				$2\textrm{tr}(\partial\mathbf{B}_d)\textrm{tr}(\mathbf{B}) + 
				4\textrm{tr}(\partial\bm{\Gamma}_d\textrm{tr}(\bm{\Gamma}))$\\
				\texttt{Var1}            & $\|\Esp{\resp{}{\omega}}{\An\odot\An}\|_F^2$ & 
				$\|\mathbf{B}\odot\mathbf{B}+ 2 \bm{\Gamma}\odot\bm{\Gamma}\|_F^2$ &  
				$2(2\mathbf{B}\odot\partial\mathbf{B}_d + 4 
				\bm{\Gamma}\odot\partial\bm{\Gamma}_d) \|\mathbf{B}\odot\mathbf{B}+ 2 
				\bm{\Gamma}\odot\bm{\Gamma}|\|_F^2$\\
				\texttt{Var2}           & $\|\Esp{\resp{}{\omega}}{\An\An}\|_F^2$ & 
				$\|\mathbf{B}\mathbf{B} + 2 
				\bm{\Gamma}\bm{\Gamma}\|_F^2$ & $2(\mathbf{B}\partial\mathbf{B}_d + 
				\partial\mathbf{B}_d\mathbf{B} + 2\bm{\Gamma}\partial\bm{\Gamma}_d + 
				2\partial\bm{\Gamma}_d\bm{\Gamma})\|\mathbf{B}\mathbf{B} + 2 
				\bm{\Gamma}\bm{\Gamma}\|_F^2$\\
			\end{tabular}
		}
	\end{table}

    The details are given in the Appendix, where gradients are also discussed.
    A brief summary of the approach is given in Algorithm \ref{alg:seq}. \resp{2-14}{The optimization with respect to $\xnew$ is non-convex; we pursue a multi-start approach by evaluating the acquisition function on a space filling design, then executing local searches via L-BFGS-B \citep{Morales2011} on the five points with the lowest acquisition function value.}

    \begin{algorithm}[ht]
	\caption{Pseudocode for the sequential learning approach}
	\begin{algorithmic}[1]
	\Require $n_0$, criteria $J(\cdot)$ (e.g., \texttt{Trace}, \texttt{Var1}, 
	\texttt{Var2})
	\State Construct and evaluate an initial design of experiments of size $n_0$ in 
	$\Xset$
	\State Build initial GP model
	\While{time/evaluation budget not exhausted}
	\State Find $\xnew^* \in \arg \max_{\x \in \Xset} J(\x)$
	\State Evaluate $f(\xnew^*)$
	\State Update the GP model based on new data
	\EndWhile
	\end{algorithmic}
	\label{alg:seq}
	\end{algorithm}

    We now turn to application examples to demonstrate the benefits of
    sequential design for active subspace learning.

	\section{Numerical Experiments}
	\label{sec:experiments}
	
    We illustrate the advantage of judicious design-point selection by
    comparing sequential GP subspace estimation with random GP estimation and
    Monte Carlo estimation, as well as approaches using a global linear model
    (OLS) or local linear models (LL). The task at hand being estimation of a
    subspace, we measure error in a model's prediction as the sine of the
    first principal angle between the true active subspace and the active
    subspace estimated at each potentially expensive function evaluation. The
    sine of the first principal angle between two subspaces $\mathcal{U}$ and
    $\hat{\mathcal{U}}$, denoted as $\angle_1 {\mathcal{U}
    \hat{\mathcal{U}}}$, may be computed as the spectral norm of a simple
    expression of two unitary matrices, the first, $\mathbf{U}$, with range
    equal to $\mathcal{U}$ and the second, $\hat{\mathbf{U}}^\perp$, with
    kernel equal to $\hat{\mathcal{U}}$ \citep[Lemma 3.1]{Jiguang87}:
	\begin{equation}
	\label{eq:subspacemetric}
			 \sin\left(\angle_1 {\mathcal{U} \hat{\mathcal{U}}} \right) = 
\left\|\mathbf{U}^\top\hat{\mathbf{U}}^\perp\right\|_2.
	\end{equation}
	This ``subspace distance'' gives us some notion of the largest possible angle to be
	made between subspaces.
	
	All of the tested methods begin with the same
 	Latin hypercube sample and choose their next points either randomly or
 	based on an acquisition function.

	\subsection{The Rank-1 Quadratic}
	
\begin{figure}[htpb]
		\centering
		\newcommand{\natesscale}{0.39}
		\includegraphics[scale=\natesscale]{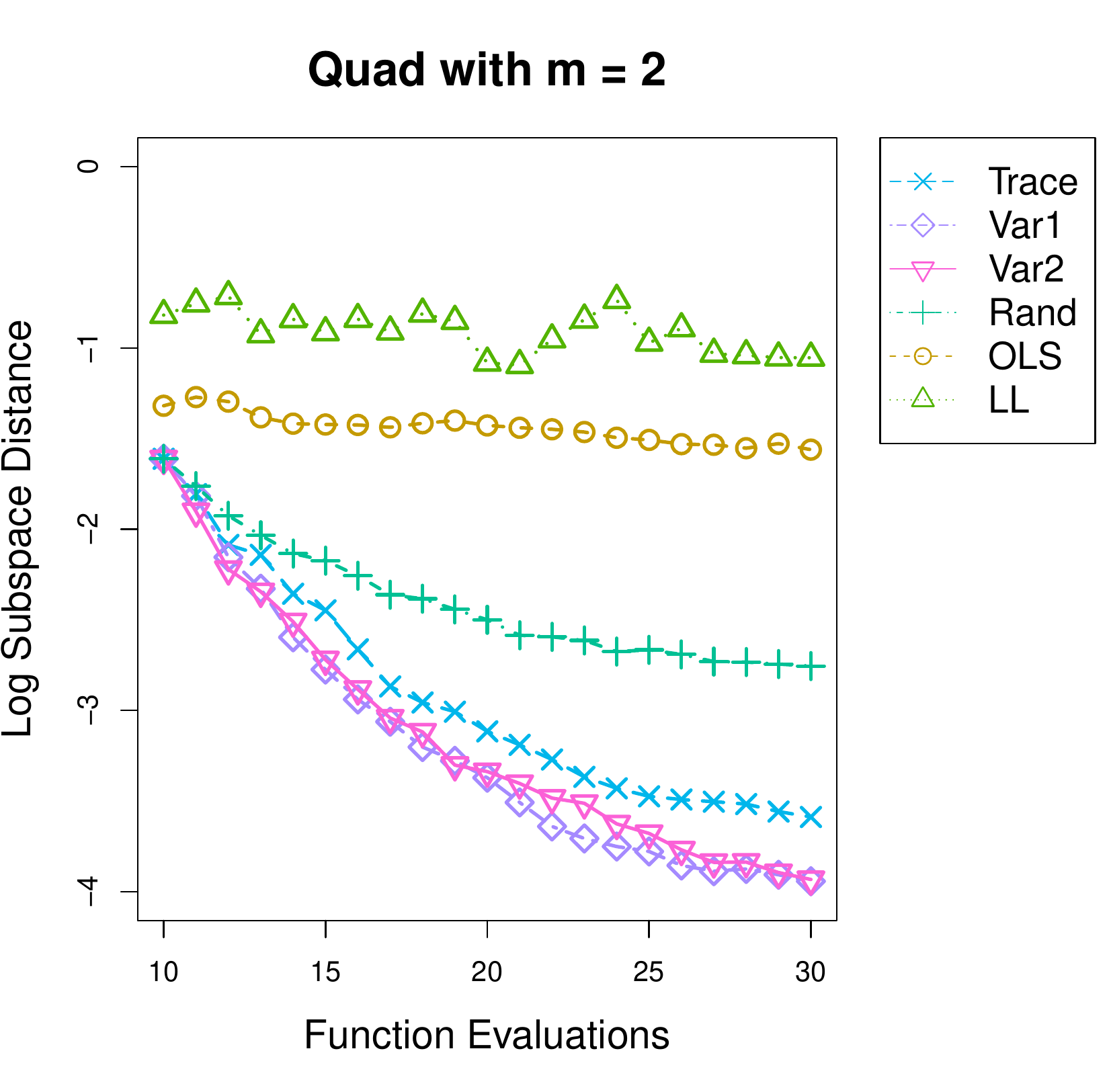}
		\includegraphics[scale=\natesscale]{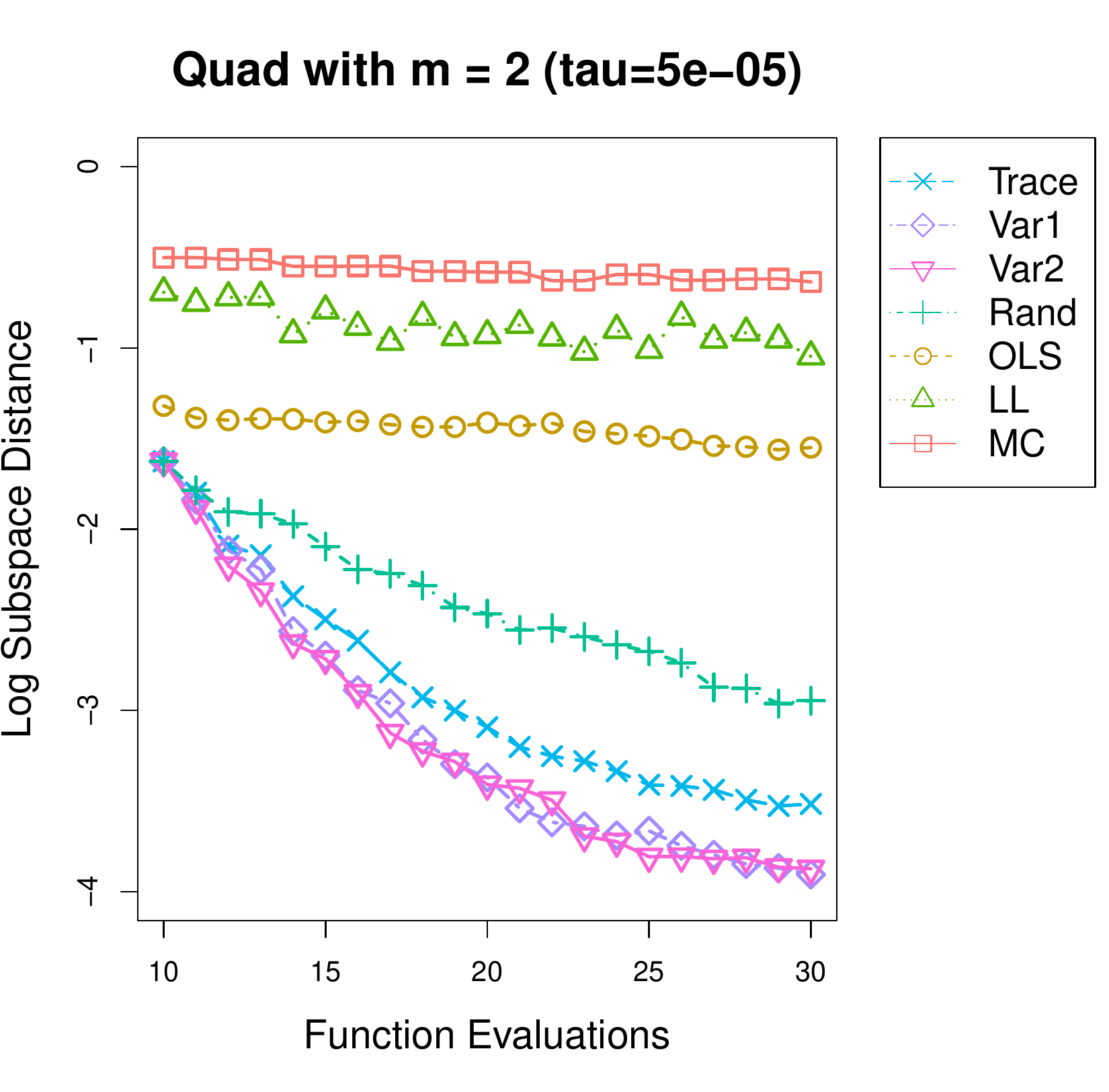}
		
		\includegraphics[scale=\natesscale]{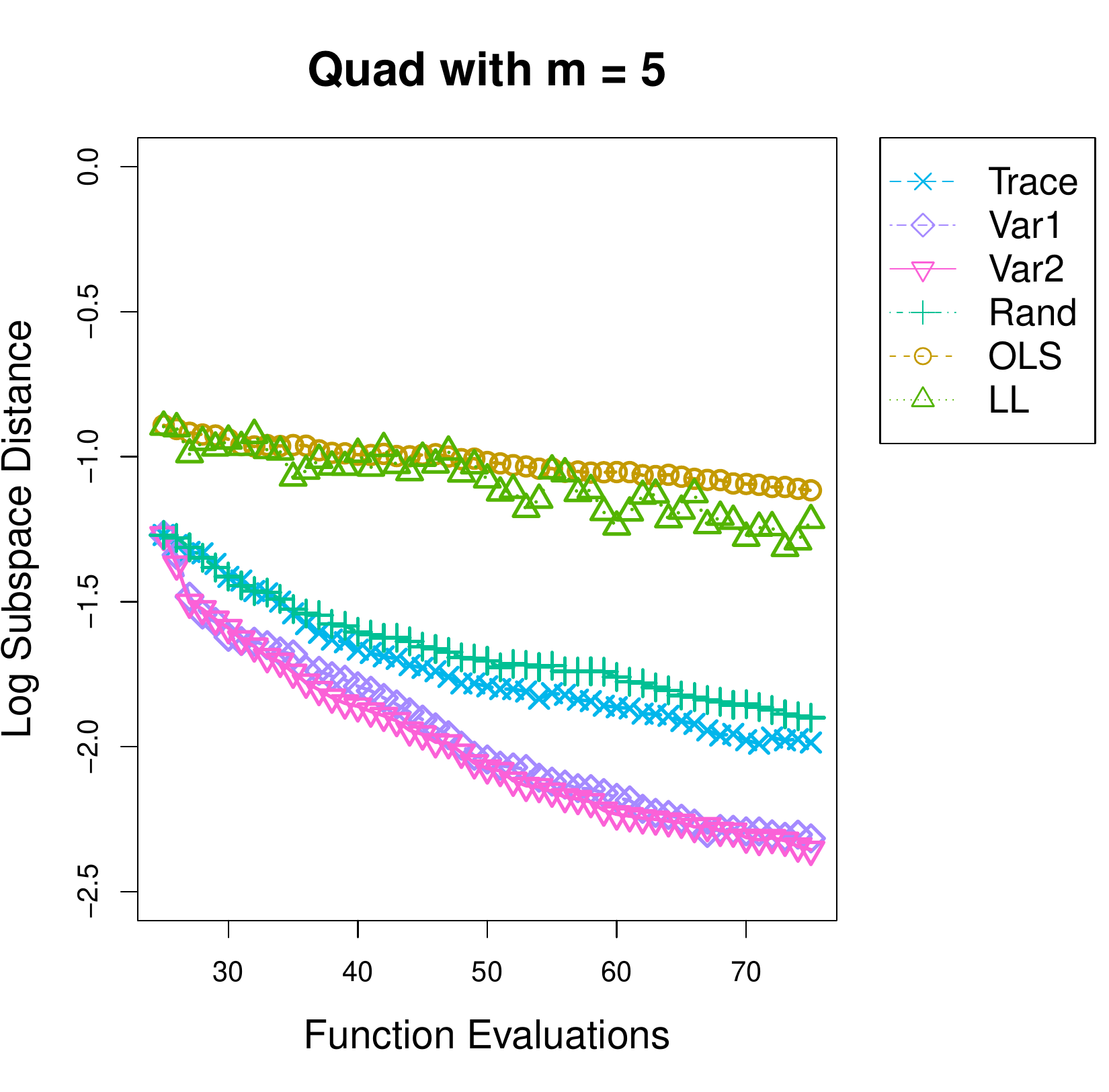}
		\includegraphics[scale=\natesscale]{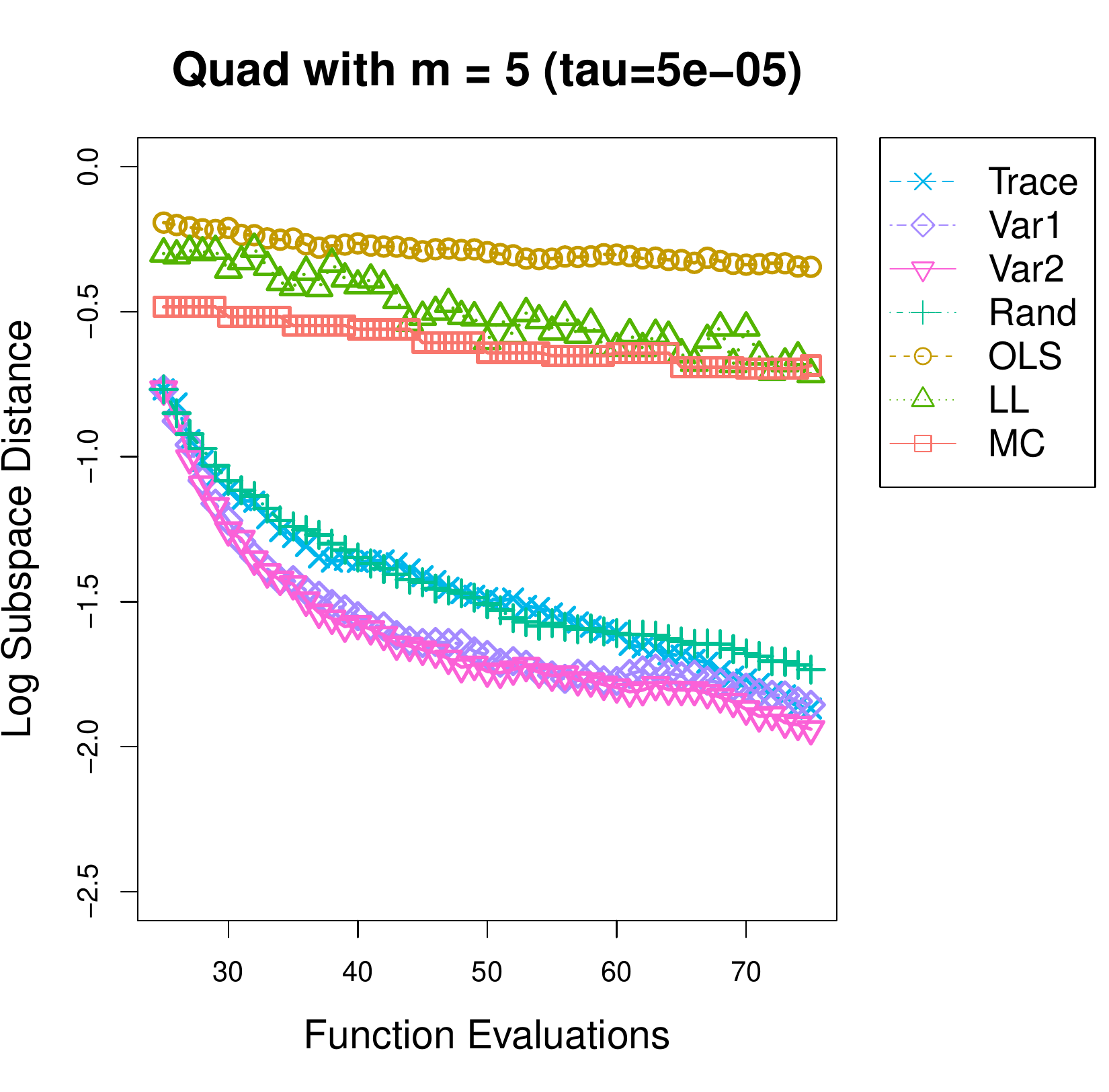}
		
		\includegraphics[scale=\natesscale]{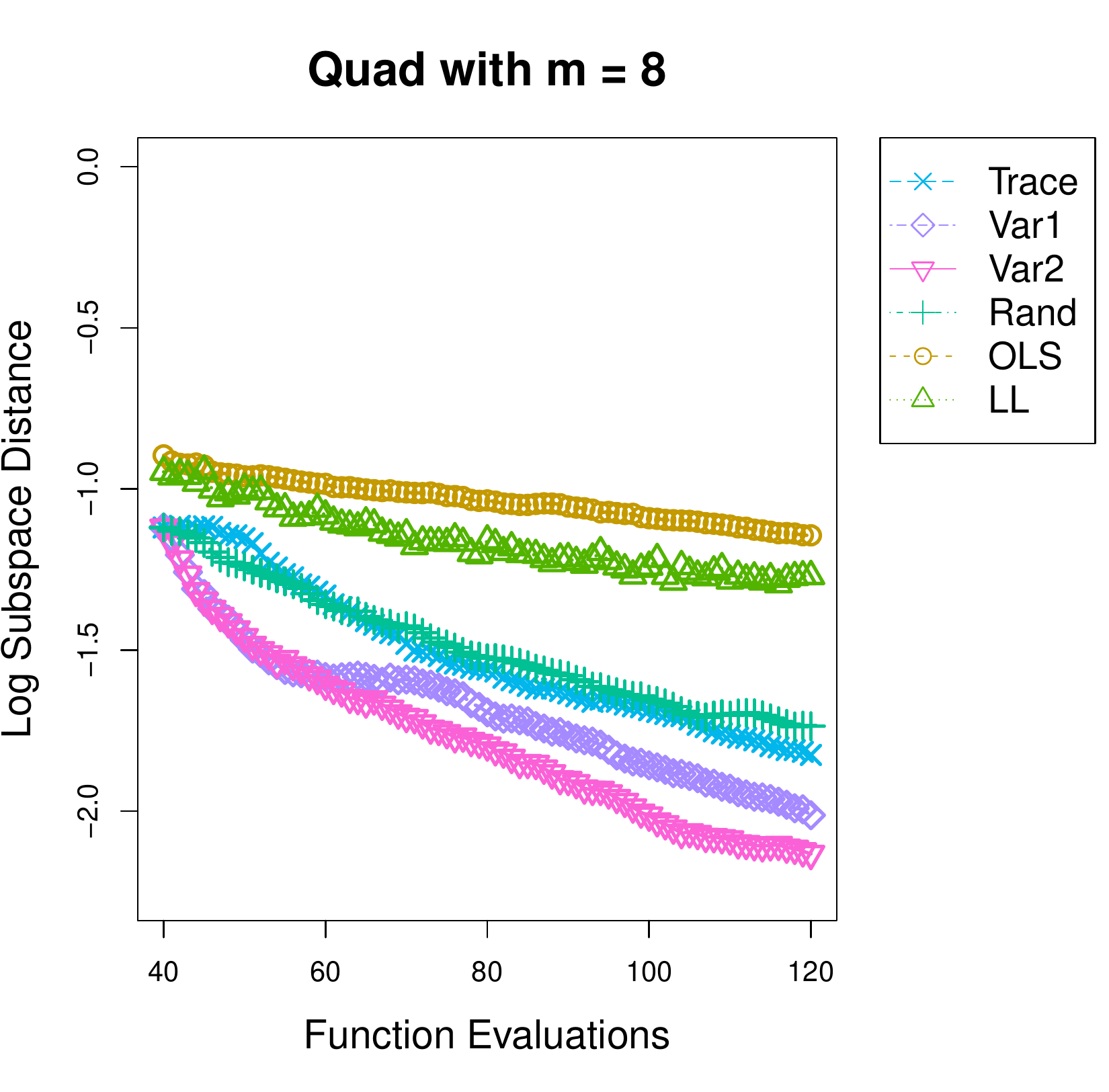}
		\includegraphics[scale=\natesscale]{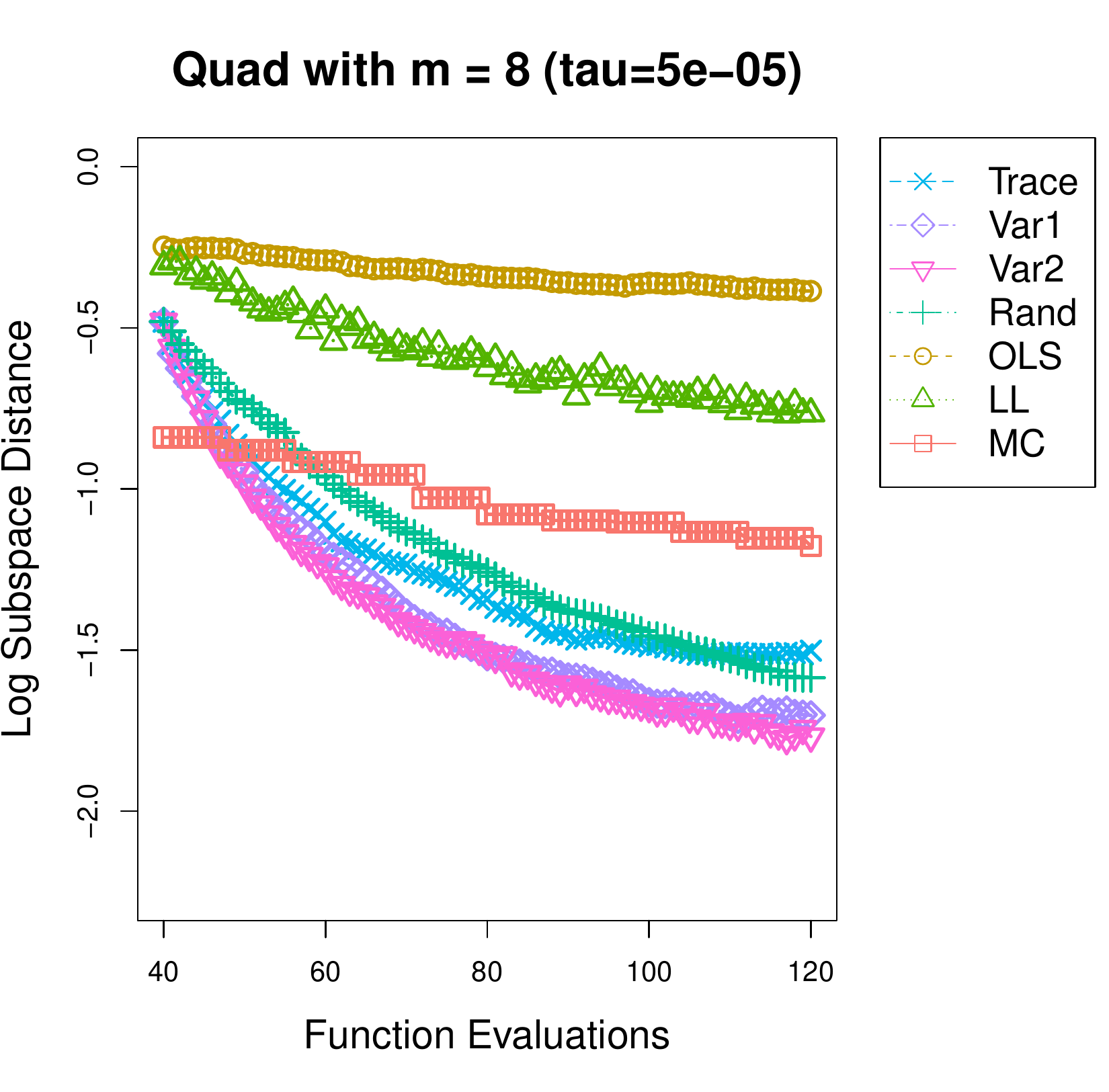}
		
		\caption{\small The quadratic problem \eqref{eq:rank1quad} in
		dimensions 2, 5, and 8. The $x$-axis represents computational effort
		in function evaluations while the $y$-axis represents the log of the mean (based on 50 trials) of the subspace error.
	 	The standard Monte Carlo estimator has zero error
		after a single gradient evaluation modulo minimal numerical noise in
		the deterministic case, so is not pictured in column 1, but would be at
		$-\infty$ in theory and was found to be around $10^{-10}$ in practice.}
		\label{fig:quad}
	\end{figure}

	We begin by learning a quadratic function defined by a rank-1 matrix
	$\mathbf{A}$; that is,

	\begin{equation}
	\label{eq:rank1quad}
		f(\x) = \x^\top \mathbf{A} \x = \left(\mathbf{a}_1^\top\x\right)^2, \mathbf{x} \in [0,1]^\nvar
	\end{equation}
    The gradient of this function is $\nabla f(\x) = 2\mathbf{A}\x = 2
    \left(\mathbf{a}_1^\top\x\right) \mathbf{a}_1$, so gradients live entirely
    in the one-dimensional subspace given by the range of $\mathbf{A}$.
    During each trial, a vector $\mathbf{a}_1$ is generated with iid standard normal entries.
    The results of applying each algorithm to 50 randomly generated rank-1
    $\mathbf{A}$ matrices
    in dimensions $\nvar=$ 2, 5, and 8 are given in Figure~\ref{fig:quad}. The
    initial design is 5 times the dimension of the space, and the number of
    points selected sequentially is 10 times the dimension, giving a total
    budget of 15 times the input dimension in terms of function evaluations.
    To simulate numerical error or a stochastic experiment, we rerun the
    experiments this time adding Gaussian noise to the objective with standard deviation 
    $\tau = 5e^{-5}$. In the
    noiseless case, as soon as we see one gradient, we have the active
    subspace, so finite differencing (simple forward differencing with a step size of $10^{-4}$) together with the Monte Carlo estimator
    gets the right answer after $\nvar+1$ evaluations. With slight numerical noise, however, 
    the accuracy obtained by finite differencing degrades, and the
    advantage of intelligent selection of evaluation points is clear.

	The difference between the selected three definitions of variance of
	$\CGPnew$ can be observed in relation to the random infill, that is,
	randomly choosing the next point. \texttt{Var1} and \texttt{Var2} give
	similar results; \texttt{Trace} is in some cases barely better than random
	sampling. This indicates that to pinpoint the active subspace, it is insufficient to
	focus on the uncertainty associated with the mean (or sum) of the
	eigenvalues. Recall that each method begins with the exact same design, and thus that 
	discrepancies in accuracy at the beginning of the simulation are due to 
	how the methods differ in exploiting the same available data. Since the 
	classical MC estimator cannot avail itself of random function evaluations,
	it is given that many additional function evaluations to conduct finite differencing
	prior to the trials that appear in the figures.
	
	\subsection{The Wing Weight Function}
	
\begin{figure}[ht]
		\centering
		\includegraphics[scale=0.45]{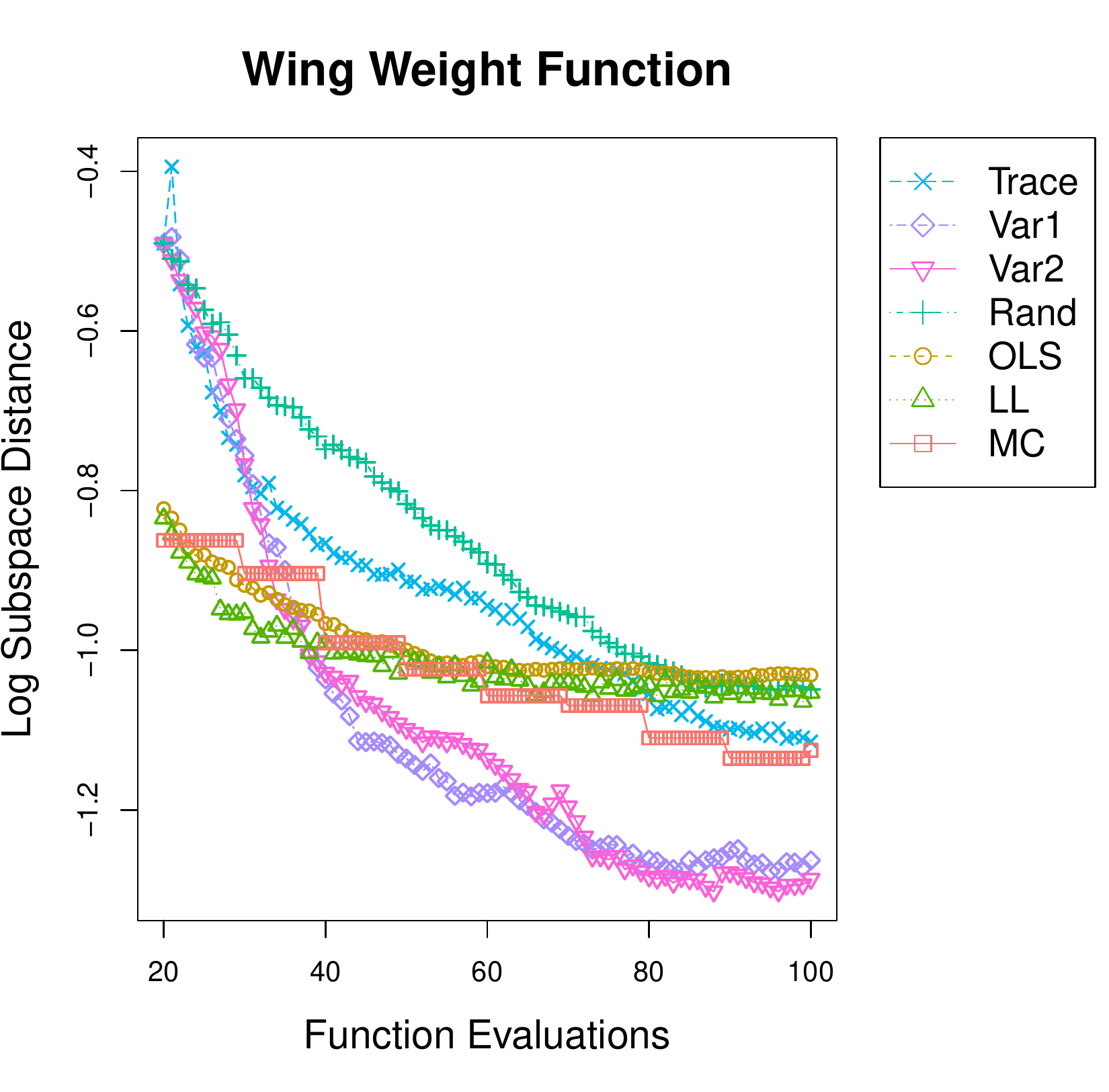}
		\includegraphics[scale=0.45,trim={0 0 3.8cm 0},clip]{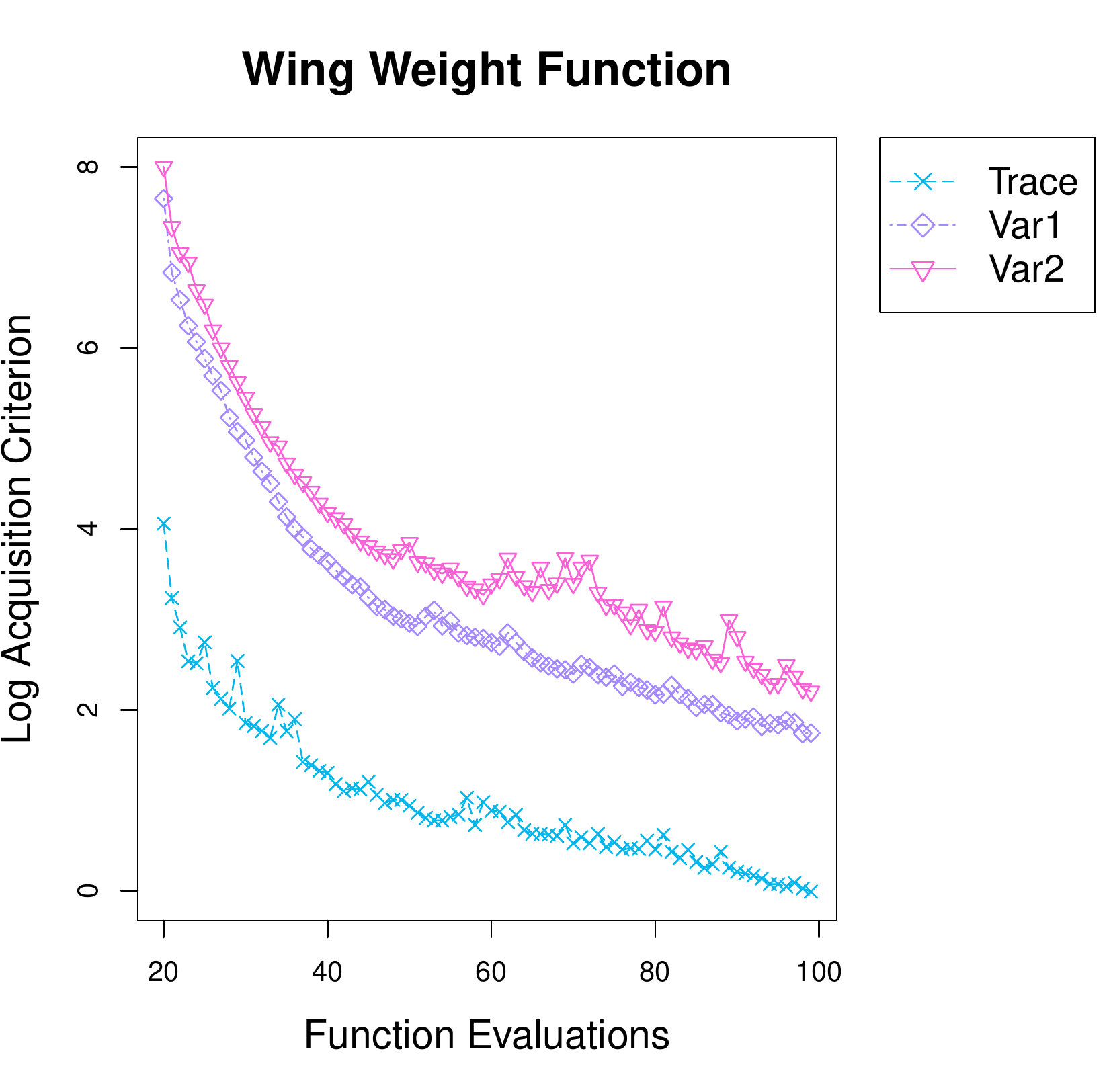}
		\caption{Results on a simulation on the wing weight function. Left:
		Function evaluations are represented with the $x$-axis, while subspace
		error at each iteration is shown on the $y$-axis. The advantage of
		sequential design for smaller function evaluation budgets is clear.
		Right: The value maximizing the
		acquisition function is given for sequential strategies at each
		iteration.}
		\label{fig:wing_weight}
	\end{figure}

\begin{figure}[ht]
		\centering
		\includegraphics[scale=0.45]{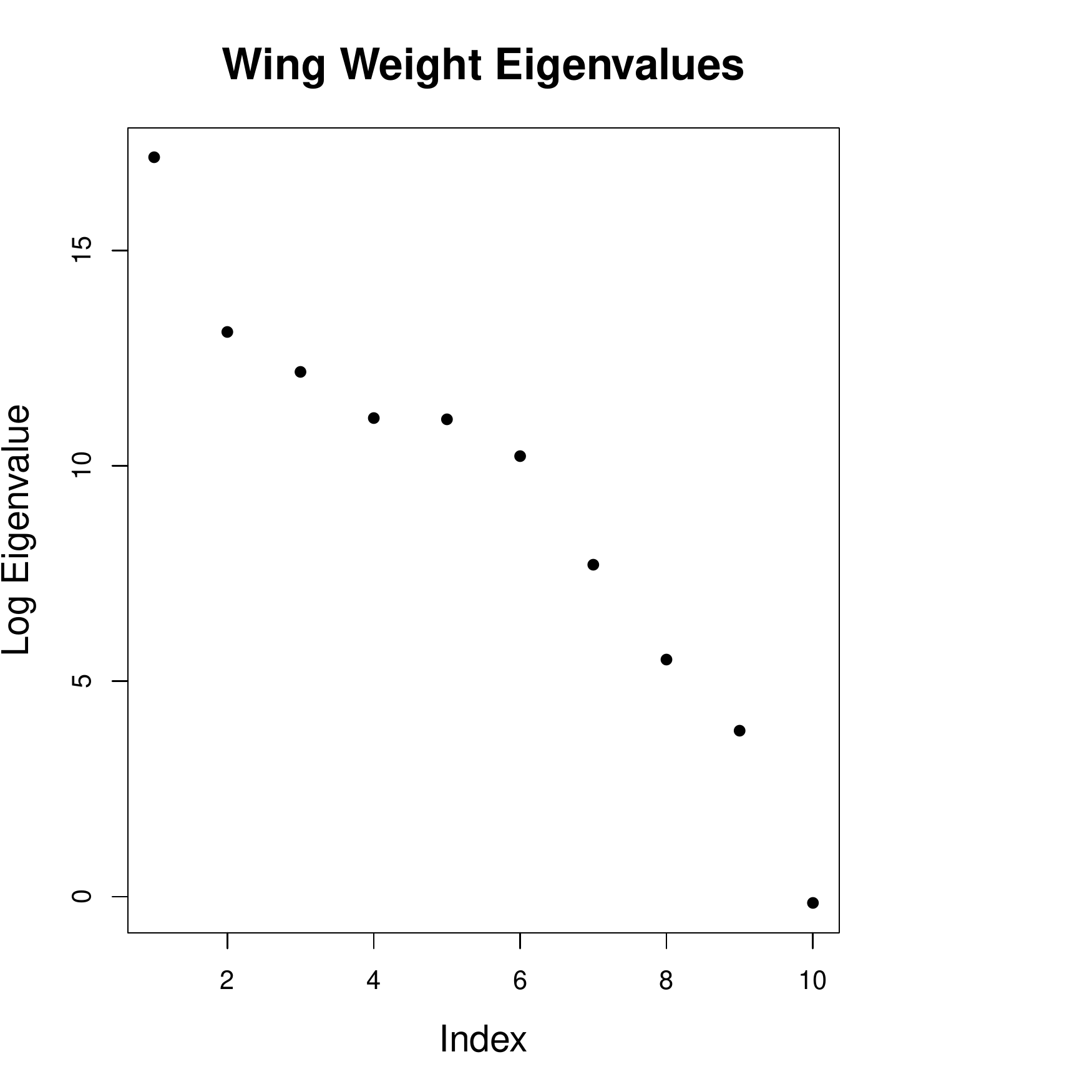}
		\includegraphics[scale=0.45]{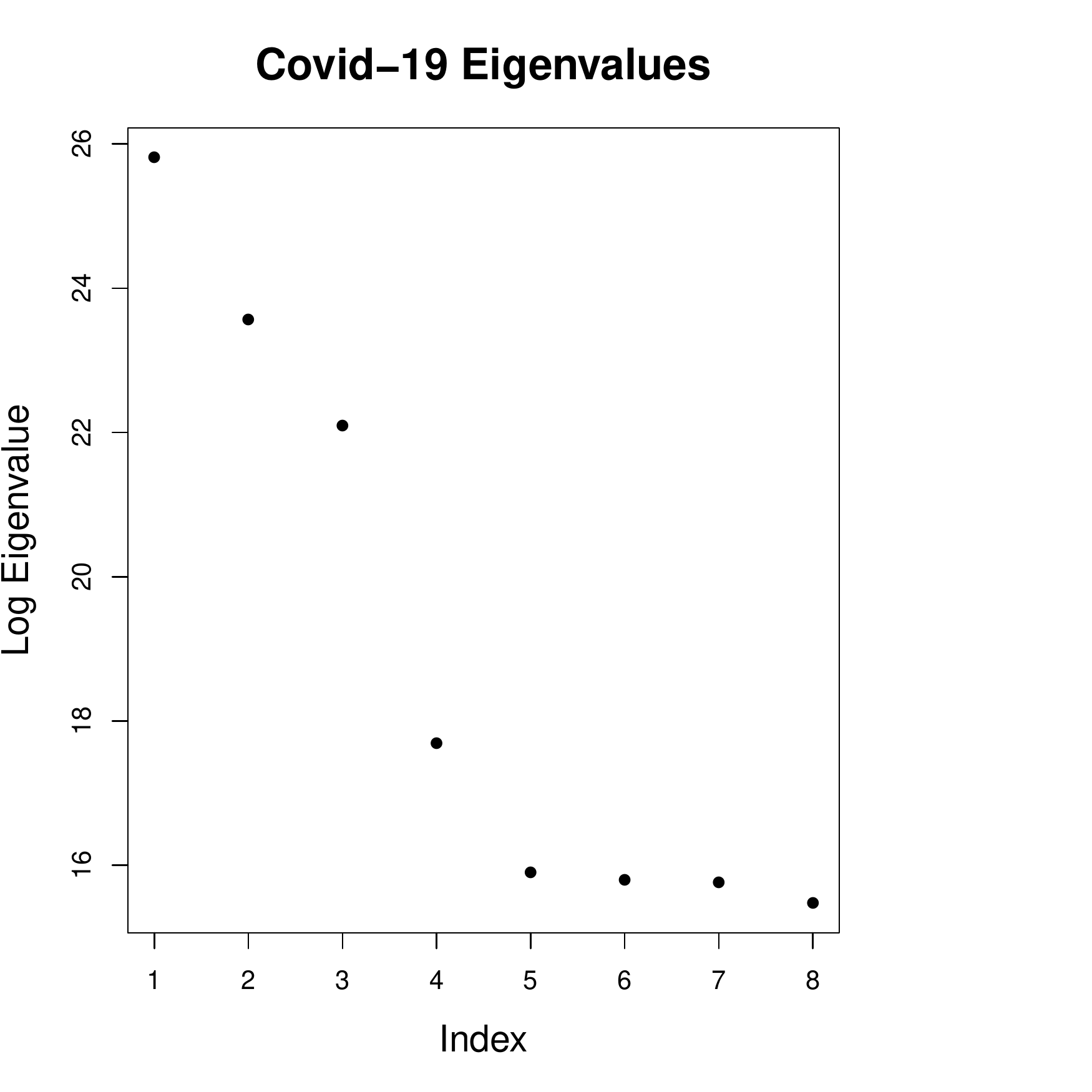}
		\caption{\resp{}{The log of the eigenvalues of $\C$ matrices estimated with many data. Gaps indicate potential cut-offs for the active subspace. The Wing Weight active subspace matrix was estimated via Finite Differencing on 10,000 random points. The Covid-19 active subspace matrix was estimated via our GP methodology with 2,000 points, as the simulator's low accuracy precludes finite differencing. We chose active subspace dimensions of 1 and 3, respectively, as referenced in the experiments.}}
		\label{fig:eigvals}
	\end{figure}

    The wing weight function, introduced by \citet{Forrester2011}, is a
    function of 10 inputs on a rectangular domain giving the weight of a light
    aircraft wing. Although it is a simple algebraic expression, it is of
    interest as a physically motivated test problem. Previously examined by
    the tutorial for the \texttt{active\_subspaces} \texttt{Python}
    package\footnote{\url{https://github.com/paulcon/active_subspaces/blob/master/tutorials/basic.ipynb}},
    a prominent one-dimensional active subspace was discovered \resp{1-1}{(see Figure \ref{fig:eigvals})}. Here, we
    explore this function beginning with a design of 20 points and choose an
    additional 80 points via a sequential design policy or randomly. Since an
    analytic representation of the active subspace is unavailable, we measure
    the ``true" active subspace by using finite differencing with the
    classical estimation technique with 10,000 function evaluations.
    Figure~\ref{fig:wing_weight} illustrates the advantage of design-point
    selection via our acquisition criteria for smaller function evaluation
    budgets. The variances that account for non-axis-aligned terms of
    $\CGPnew$ (i.e., in contrast to the \texttt{Trace} acquisition    
    function) perform best.
	
	\resp{1-2}{
	\subsection{The Minnesota Covid-19 Simulator}
	
	The Minnesota Covid-19 Model \citep{mncovid} was developed to help guide public policy in response to the novel coronavirus, and calculates trajectories of various quantities of interest, such as cumulative death rates and ICU utilization over one year. However, the model depends on various epidemiological quantities, such as rate of asymptomatic infection, which are presently poorly understood. In this section, we treat this simulator as a black box and study the effect of varying certain parameters
	\footnote{Namely, the number of initial cases, the contact reduction of shelter in place and social distancing countermeasures, the proportion by which contact is reduced in persons aged 60 or more, the contact reduction due to personal behavior changes, the proportion of asymptomatic cases, the probability of a person over 80 being hospitalized, and the probability of a person over 80 dying at home}
	 on the total covid-19 deaths. The cumulative number of deaths, though calculated as a continuous variable, is not sufficiently precise so as to allow finite differencing. Thus, in our simulations, accuracy is computed as compared to the active subspace given by a Gaussian process fit on 2,000 randomly sampled data (Figure \ref{fig:eigvals} reveals that a three dimensional active subspace is reasonable). Figure \ref{fig:covid} illustrates the performance of our three sequential design criteria as compared to a random baseline. Again the variance-based acquisition functions perform best, while the \texttt{Trace} function is in this case only marginally better than random.
	
		\begin{figure}[ht]
		\centering
		\includegraphics[scale=0.45]{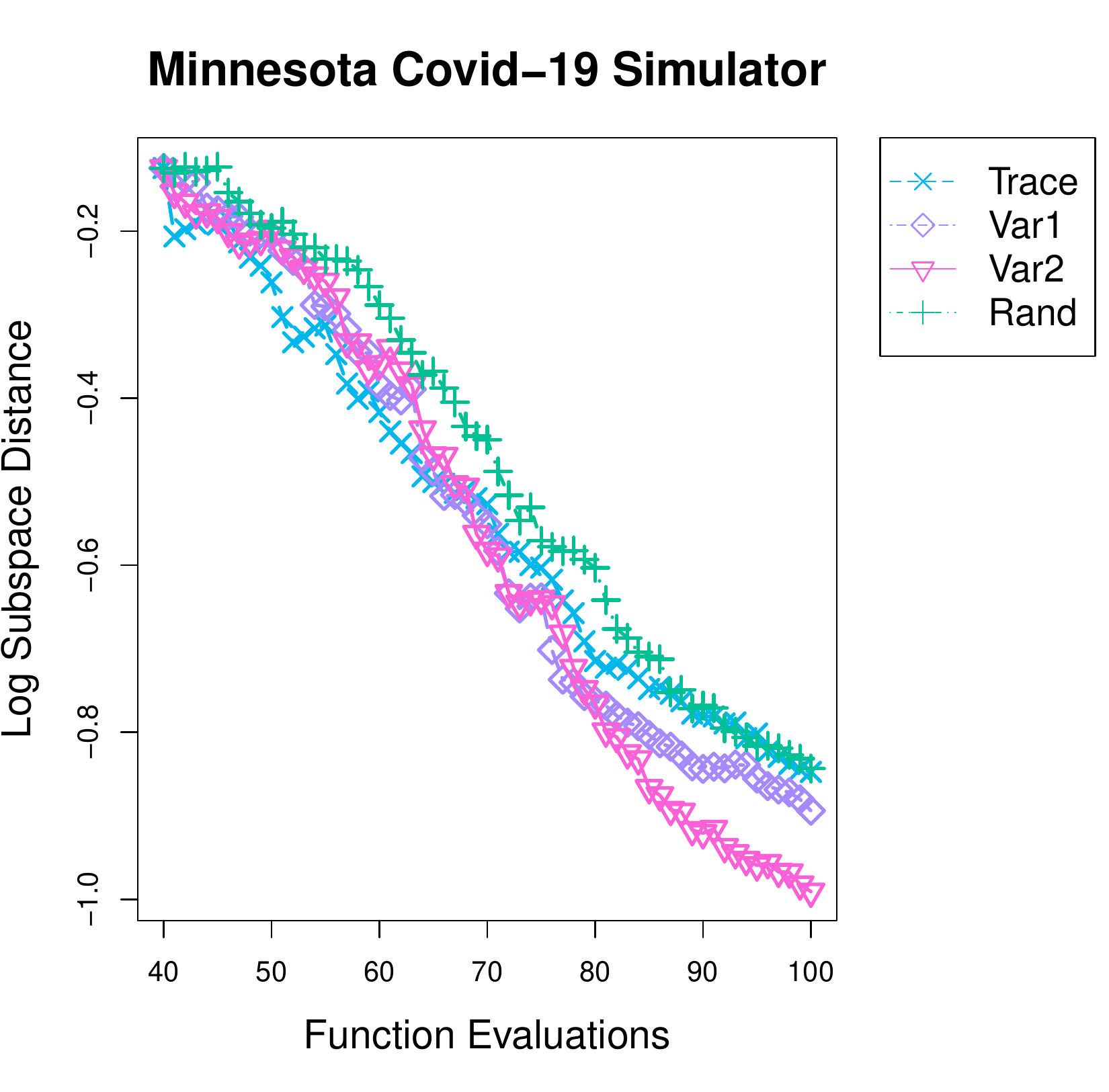}
		\includegraphics[scale=0.45]{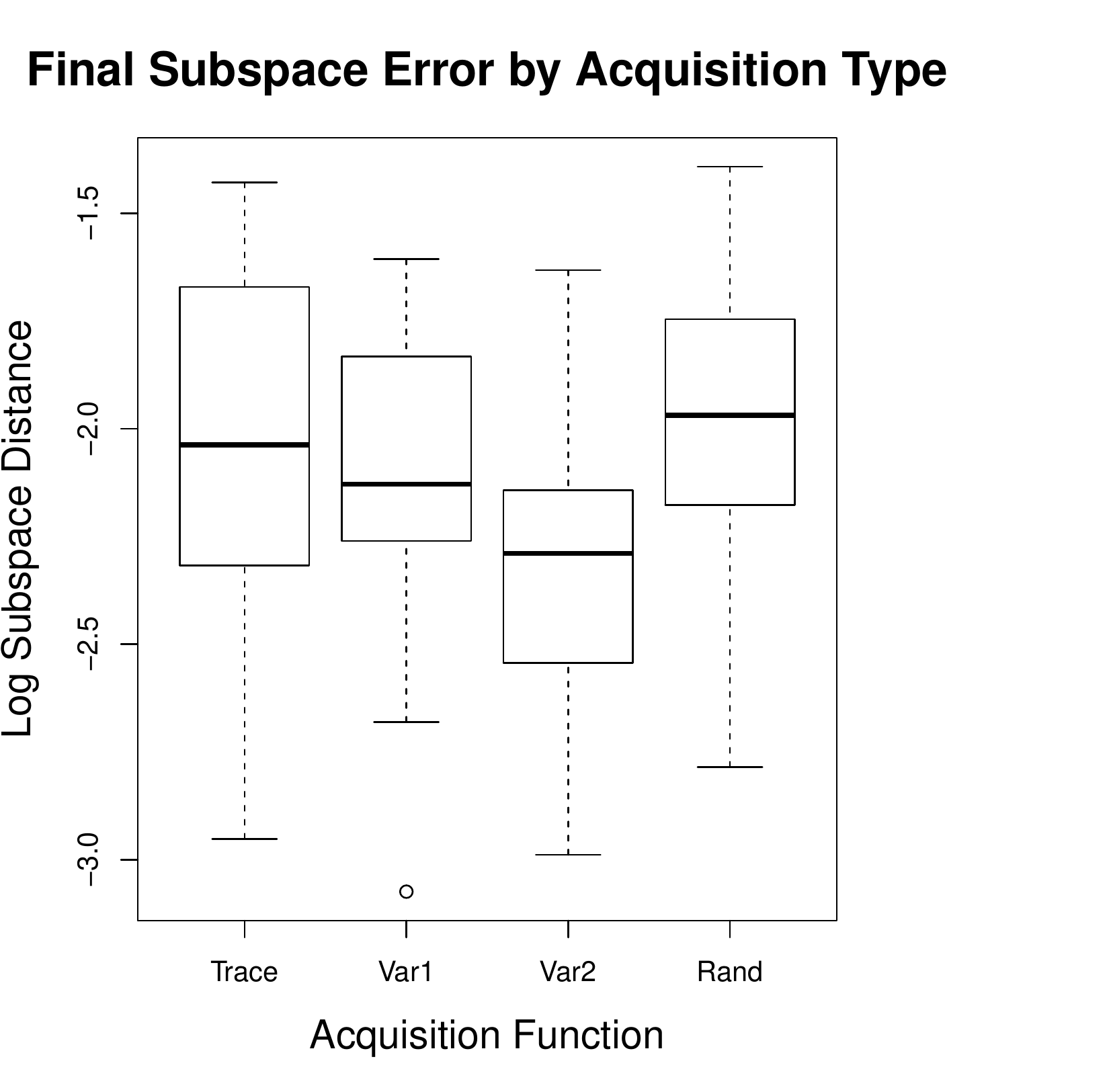}
		\caption{\resp{}{Comparison of GP based sequential design criteria on the Minnesota Covid-19 Simulator. Ground truth is taken to be the active subspace given by a GP with 2,000 randomly sampled input-output pairs. Left: Subspace error versus function evaluations, averaged over 40 runs. Right: Boxplots showing the final subspace error for each acquisition function over each of the 40 runs.}}
		\label{fig:covid}
	\end{figure}
}
	\section{Discussion and Perspectives}
	\label{sec:discussion}
	In this article, we have shown that the surrogate-assisted estimation of
	active subspaces, a technique previously executed via Monte Carlo finite
	differencing, has a closed-form solution. This result was leveraged to
	create acquisition functions enabling evaluation-efficient estimation of
	such subspaces via sequential design, potentially saving (expensive)
	black-box evaluations. Although in this article we initialized our
	search with a standard Latin hypercube sample, choosing an initial design
	related to our acquisition criteria may be preferable, an issue we leave
	for future work. Active subspace methods being a field of active research,
	we leave the estimation via Gaussian processes of recent refinements, such
	as that proposed by \cite{Lee2019}, for perspective. Future work could
	focus on problems specifically related to optimization, perhaps by
	considering a different measure to define the expected outer product of
	the gradient or by explicitly incorporating only those subspaces that play
	an important role in optimization, perhaps phrased in the framework of
	goal-oriented sensitivity analysis \citep{Fort2016}. Although not explored
	in this work, it is in principle straightforward to choose which points in
	the design space to explore in batches rather than one at a time. Such a
	problem, however, presents computational difficulty in optimization of the
	acquisition function. One may have to develop approximate methods in order
	for this to be carried out in practice.

	\section*{Acknowledgments}
	This material was based upon work supported by the U.S.\ Department of
	Energy, Office of Science, Office of Advanced Scientific Computing
	Research, applied mathematics and SciDAC programs under Contract No.\
	DE-AC02-06CH11357. The authors would like to thank Robert B.\ Gramacy for
	thoughtful comments on early drafts. \resp{}{This article benefited greatly from feedback provided by two anonymous referees.}
    
	\bibliography{sasl}

\begin{thebibliography}{53}
\newcommand{\enquote}[1]{``#1''}
\expandafter\ifx\csname natexlab\endcsname\relax\def\natexlab#1{#1}\fi

\bibitem[\protect\citename{Barnett, }1979]{barnett:1979}
Barnett, S. (1979).
\newblock {\em Matrix Methods for Engineers and Scientists\/}.
\newblock McGraw-Hill.

\bibitem[\protect\citename{Bellman, }2003]{Bellman2003}
Bellman, R.~E. (2003).
\newblock {\em Dynamic Programming\/}.
\newblock New York, NY: Dover Publications.

\bibitem[\protect\citename{Binois et~al., }2019]{Binois2018}
Binois, M., Huang, J., Gramacy, R.~B., and Ludkovski, M. (2019).
\newblock \enquote{Replication or Exploration? Sequential Design for Stochastic
  Simulation Experiments.}
\newblock {\em Technometrics\/}, 61, 1, 7--23.

\bibitem[\protect\citename{Constantine, }2015]{Constantine2015}
Constantine, P.~G. (2015).
\newblock {\em Active Subspaces\/}.
\newblock Philadelphia, PA: SIAM.

\bibitem[\protect\citename{Constantine et~al., }2014]{Constantine2014}
Constantine, P.~G., Dow, E., and Wang, Q. (2014).
\newblock \enquote{Active Subspace Methods in Theory and Practice: Applications
  to Kriging Surfaces.}
\newblock {\em SIAM Journal on Scientific Computing\/}, 36, 4, A1500--A1524.

\bibitem[\protect\citename{Constantine et~al.,
  }2017]{constantine2017stationary}
Constantine, P.~G., Eftekhari, A., Hokanson, J., and Ward, R.~A. (2017).
\newblock \enquote{A Near-Stationary Subspace for Ridge Approximation.}
\newblock {\em Computer Methods in Applied Mechanics and Engineering\/}, 326,
  402--421.

\bibitem[\protect\citename{Constantine et~al., }2015]{constantine2015sketch}
Constantine, P.~G., Eftekhari, A., and Wakin, M.~B. (2015).
\newblock \enquote{Computing Active Subspaces Efficiently with Gradient
  Sketching.}
\newblock In {\em 2015 IEEE 6th International Workshop on Computational
  Advances in Multi-Sensor Adaptive Processing (CAMSAP)\/},  353--356. IEEE.

\bibitem[\protect\citename{De~Lozzo and Marrel, }2016]{DeLozzo2016}
De~Lozzo, M. and Marrel, A. (2016).
\newblock \enquote{Estimation of the Derivative-Based Global Sensitivity
  Measures Using a {Gaussian} Process Metamodel.}
\newblock {\em SIAM/ASA Journal on Uncertainty Quantification\/}, 4, 1,
  708--738.

\bibitem[\protect\citename{Djolonga et~al., }2013]{Djolonga2013}
Djolonga, J., Krause, A., and Cevher, V. (2013).
\newblock \enquote{High-Dimensional {Gaussian} Process Bandits.}
\newblock In {\em Advances in Neural Information Processing Systems 26\/},
  1025--1033.

\bibitem[\protect\citename{Durrande et~al., }2012]{Durrande2010}
Durrande, N., Ginsbourger, D., and Roustant, O. (2012).
\newblock \enquote{Additive Kernels for {Gaussian} Process Modeling.}
\newblock {\em Annales de la Facult{\'e}e de Sciences de Toulouse\/}, ~17.

\bibitem[\protect\citename{Durrande et~al., }2013]{Durrande2013}
Durrande, N., Ginsbourger, D., Roustant, O., and Carraro, L. (2013).
\newblock \enquote{{ANOVA} Kernels and {RKHS} of Zero Mean Functions for
  Model-Based Sensitivity Analysis.}
\newblock {\em Journal of Multivariate Analysis\/}, 115, 57--67.

\bibitem[\protect\citename{Duvenaud et~al., }2011]{Duvenaud2011}
Duvenaud, D.~K., Nickisch, H., and Rasmussen, C.~E. (2011).
\newblock \enquote{Additive {Gaussian} Processes.}
\newblock In {\em Advances in Neural Information Processing Systems 24\/}, eds.
  J.~Shawe-Taylor, R.~S. Zemel, P.~L. Bartlett, F.~Pereira, and K.~Q.
  Weinberger,  226--234. Curran Associates, Inc.

\bibitem[\protect\citename{Efron, }1981]{Efron1981}
Efron, B. (1981).
\newblock \enquote{Nonparametric Estimates of Standard Error: The Jackknife,
  the Bootstrap and Other Methods.}
\newblock {\em Biometrika\/}, 68, 3, 589--599.

\bibitem[\protect\citename{Enns et~al., }2020]{mncovid}
Enns, E.~A., Kirkeide, M., Mehta, A., MacLehose, R., Knowlton, G.~S., Smith,
  M.~K., Searle, K.~M., Zhao, R., Gildemeister, S., Simon, A., Sanstead, E.,
  and Kulasingam, S. (2020).
\newblock \enquote{Modeling the Impact of Social Distancing Measures on the
  Spread of {SARS-CoV-2} in {Minnesota}.}
\newblock Tech. Rep. 1148-427724, University of Minnesota.

\bibitem[\protect\citename{Forrester et~al., }2008]{Forrester2011}
Forrester, A. I.~J., Sobester, A., and Keane, A.~J. (2008).
\newblock {\em Engineering Design via Surrogate Modelling - A Practical
  Guide\/}.
\newblock Wiley.

\bibitem[\protect\citename{Fort et~al., }2016]{Fort2016}
Fort, J.-C., Klein, T., and Rachdi, N. (2016).
\newblock \enquote{New Sensitivity Analysis Subordinated to a Contrast.}
\newblock {\em Communications in Statistics - Theory and Methods\/}, 45, 15,
  4349--4364.

\bibitem[\protect\citename{Fr\'echet, }1948]{frechet1948}
Fr\'echet, M.~R. (1948).
\newblock \enquote{Les \'El\'ements Al\'eatoires de Nature Quelconque dans un
  Espace Distanci\'e.}
\newblock {\em Annales de l'Institut Henri Poincar\'e\/}, 10, 4, 215--310.

\bibitem[\protect\citename{Friedman and Stuetzle, }1981]{Friedman1981}
Friedman, J.~H. and Stuetzle, W. (1981).
\newblock \enquote{Projection Pursuit Regression.}
\newblock {\em Journal of the American Statistical Association\/}, 76, 376,
  817--823.

\bibitem[\protect\citename{Fukumizu and Leng, }2014]{Fukumizu2014}
Fukumizu, K. and Leng, C. (2014).
\newblock \enquote{Gradient-Based Kernel Dimension Reduction for Regression.}
\newblock {\em Journal of the American Statistical Association\/}, 109, 505,
  359--370.

\bibitem[\protect\citename{Garnett et~al., }2014]{Garnett2014}
Garnett, R., Osborne, M.~A., and Hennig, P. (2014).
\newblock \enquote{Active Learning of Linear Embeddings for {Gaussian}
  Processes.}
\newblock In {\em Proceedings of the Thirtieth Conference on Uncertainty in
  Artificial Intelligence\/}, UAI'14,  230--239. AUAI Press.

\bibitem[\protect\citename{Ghosh, }2018]{Ghosh2018}
Ghosh, S. (2018).
\newblock {\em Kernel Smoothing: Principles, Methods and Applications\/}.
\newblock John Wiley \& Sons.

\bibitem[\protect\citename{Glaws et~al., }2020]{Glaws2020}
Glaws, A., Constantine, P.~G., and Cook, R.~D. (2020).
\newblock \enquote{Inverse regression for ridge recovery: a data-driven
  approach for parameter reduction in computer experiments.}
\newblock {\em Statistics and Computing\/}, 30, 2, 237--253.

\bibitem[\protect\citename{Hokanson and Constantine, }2018]{Hokanson2018}
Hokanson, J. and Constantine, P.~G. (2018).
\newblock \enquote{Data-Driven Polynomial Ridge Approximation Using Variable
  Projection.}
\newblock {\em SIAM Journal on Scientific Computing\/}, 40, 3, A1566--A1589.

\bibitem[\protect\citename{Holodnak et~al., }2018]{Holodnak2018}
Holodnak, J.~T., Ipsen, I.~C., and Smith, R.~C. (2018).
\newblock \enquote{A Probabilistic Subspace Bound with Application to Active
  Subspaces.}
\newblock {\em SIAM Journal on Matrix Analysis and Applications\/}, 39, 3,
  1208--1220.

\bibitem[\protect\citename{Iooss and Lema{\^i}tre, }2015]{Iooss2015}
Iooss, B. and Lema{\^i}tre, P. (2015).
\newblock \enquote{A Review on Global Sensitivity Analysis Methods.}
\newblock In {\em Uncertainty Management in Simulation-Optimization of Complex
  Systems: Algorithms and Applications\/}, eds. C.~Meloni and G.~Dellino,
  101--122. Springer.

\bibitem[\protect\citename{Ji-guang, }1987]{Jiguang87}
Ji-guang, S. (1987).
\newblock \enquote{Perturbation of Angles between Linear Subspaces.}
\newblock {\em Journal of Computational Mathematics\/}, 5, 1, 58--61.

\bibitem[\protect\citename{Labopin-Richard and Picheny,
  }2018]{Labopin-Richard2016}
Labopin-Richard, T. and Picheny, V. (2018).
\newblock \enquote{Sequential Design of Experiments for Estimating Percentiles
  of Black-Box Functions.}
\newblock {\em Statistica Sinica\/}, 28, 853--877.

\bibitem[\protect\citename{Larson et~al., }2019]{LMW2019AN}
Larson, J., Menickelly, M., and Wild, S.~M. (2019).
\newblock \enquote{Derivative-Free Optimization Methods.}
\newblock {\em Acta Numerica\/}, 28, 287--404.

\bibitem[\protect\citename{Lee, }2019]{Lee2019}
Lee, M.~R. (2019).
\newblock \enquote{Modified Active Subspaces Using the Average of Gradients.}
\newblock {\em SIAM/ASA Journal on Uncertainty Quantification\/}, 7, 1, 53--66.

\bibitem[\protect\citename{Li, }1991]{li1991}
Li, K.-C. (1991).
\newblock \enquote{Sliced Inverse Regression for Dimension Reduction.}
\newblock {\em Journal of the American Statistical Association\/}, 86, 414,
  316--327.

\bibitem[\protect\citename{Li, }1992]{li1992}
--- (1992).
\newblock \enquote{On Principal {Hessian} Directions for Data Visualization and
  Dimension Reduction: Another Application of {Stein's} Lemma.}
\newblock {\em Journal of the American Statistical Association\/}, 87, 420,
  1025--1039.

\bibitem[\protect\citename{Li et~al., }2016]{Li2016}
Li, W., Lin, G., and Li, B. (2016).
\newblock \enquote{Inverse regression-based uncertainty quantification
  algorithms for high-dimensional models: Theory and practice.}
\newblock {\em Journal of Computational Physics\/}, 321, 259 -- 278.

\bibitem[\protect\citename{Ma and Zhu, }2013]{yanyuan2013}
Ma, Y. and Zhu, L. (2013).
\newblock \enquote{A Review on Dimension Reduction.}
\newblock {\em International Statistical Review\/}, 81, 1, 134--150.

\bibitem[\protect\citename{Marcy, }2017]{Marcy2017}
Marcy, P.~W. (2017).
\newblock \enquote{Bayesian {Gaussian} Process Models on Spaces of Sufficient
  Dimension Reduction.}
\newblock Statistical Perspectives of Uncertainty Quantification Workshop.

\bibitem[\protect\citename{Morales and Nocedal, }2011]{Morales2011}
Morales, J.~L. and Nocedal, J. (2011).
\newblock \enquote{Remark on {\textquotedblleft}Algorithm 778: {L-BFGS-B}:
  {Fortran} Subroutines for Large-Scale Bound Constrained
  Optimization{\textquotedblright}.}
\newblock {\em {ACM} Transactions on Mathematical Software\/}, 38, 1, 1--4.

\bibitem[\protect\citename{Mor{\'e} and Wild, }2012]{More2012}
Mor{\'e}, J.~J. and Wild, S.~M. (2012).
\newblock \enquote{Estimating Derivatives of Noisy Simulations.}
\newblock {\em ACM Transactions on Mathematical Software\/}, 38, 3,
  19:1--19:21.

\bibitem[\protect\citename{Namura et~al., }2017]{Namura2017}
Namura, N., Shimoyama, K., and Obayashi, S. (2017).
\newblock \enquote{Kriging Surrogate Model with Coordinate Transformation Based
  on Likelihood and Gradient.}
\newblock {\em Journal of Global Optimization\/}, 68, 4, 827--849.

\bibitem[\protect\citename{Othmer et~al., }2016]{Othmer2016}
Othmer, C., Lukaczyk, T.~W., Constantine, P., and Alonso, J.~J. (2016).
\newblock \enquote{On Active Subspaces in Car Aerodynamics.}
\newblock In {\em 17th {AIAA}/{ISSMO} {Multidisciplinary} {Analysis} and
  {Optimization} {Conference}\/}. American Institute of Aeronautics and
  Astronautics.

\bibitem[\protect\citename{Palar and Shimoyama, }2017]{palar2017}
Palar, P.~S. and Shimoyama, K. (2017).
\newblock \enquote{Exploiting Active Subspaces in Global Optimization: How
  Complex is Your Problem?}
\newblock In {\em Proceedings of the {Genetic} and {Evolutionary} {Computation}
  {Conference} {Companion} on - {GECCO} '17\/},  1487--1494. ACM Press.

\bibitem[\protect\citename{Palar and Shimoyama, }2018]{Palar2018}
--- (2018).
\newblock \enquote{On The Accuracy of Kriging Model in Active Subspaces.}
\newblock In {\em 2018 AIAA/ASCE/AHS/ASC Structures, Structural Dynamics, and
  Materials Conference\/},  0913.

\bibitem[\protect\citename{Petersen et~al., }2008]{Petersen2008}
Petersen, K.~B., Pedersen, M.~S., et~al. (2008).
\newblock \enquote{The Matrix Cookbook.}
\newblock {\em Technical University of Denmark\/}, 7, 15.

\bibitem[\protect\citename{Rasmussen and Williams, }2006]{Rasmussen2006}
Rasmussen, C.~E. and Williams, C. (2006).
\newblock {\em Gaussian Processes for Machine Learning\/}.
\newblock MIT Press.

\bibitem[\protect\citename{Salem et~al., }2019]{Salem2018}
Salem, M.~B., Bachoc, F., Roustant, O., Gamboa, F., and Tomaso, L. (2019).
\newblock \enquote{Sequential Dimension Reduction for Learning Features of
  Expensive Black-Box Functions.}
\newblock Preprint 01688329v2, HAL.

\bibitem[\protect\citename{Samarov, }1993]{samarov1993}
Samarov, A.~M. (1993).
\newblock \enquote{Exploring Regression Structure Using Nonparametric
  Functional Estimation.}
\newblock {\em Journal of the American Statistical Association\/}, 88, 423,
  836--847.

\bibitem[\protect\citename{Scholkopf and Smola, }2001]{Scholkopf2001}
Scholkopf, B. and Smola, A.~J. (2001).
\newblock {\em Learning with Kernels: Support Vector Machines, Regularization,
  Optimization, and Beyond\/}.
\newblock Cambridge, MA: MIT Press.

\bibitem[\protect\citename{Sobol, }2001]{sobol2001}
Sobol, I. (2001).
\newblock \enquote{Global Sensitivity Indices for Nonlinear Mathematical Models
  and Their {Monte Carlo} Estimates.}
\newblock {\em Mathematics and Computers in Simulation\/}, 55, 1, 271--280.
\newblock The Second IMACS Seminar on Monte Carlo Methods.

\bibitem[\protect\citename{Sung et~al., }2019]{Sung2017}
Sung, C.-L., Wang, W., Plumlee, M., and Haaland, B. (2019).
\newblock \enquote{Multiresolution Functional {ANOVA} for Large-Scale,
  Many-Input Computer Experiments.}
\newblock {\em Journal of the American Statistical Association\/}, 115, 530,
  908--919.

\bibitem[\protect\citename{Titsias and Lawrence, }2010]{Titsias2010}
Titsias, M. and Lawrence, N.~D. (2010).
\newblock \enquote{Bayesian {Gaussian} Process Latent Variable Model.}
\newblock In {\em Proceedings of the Thirteenth International Conference on
  Artificial Intelligence and Statistics\/},  844--851.

\bibitem[\protect\citename{Tripathy et~al., }2016]{Tripathy2016}
Tripathy, R., Bilionis, I., and Gonzalez, M. (2016).
\newblock \enquote{Gaussian Processes with Built-in Dimensionality Reduction:
  Applications to High-Dimensional Uncertainty Propagation.}
\newblock {\em Journal of Computational Physics\/}, 321, 191--223.

\bibitem[\protect\citename{Viswanath et~al., }2011]{Viswanath2011}
Viswanath, A., J.~Forrester, A., and Keane, A. (2011).
\newblock \enquote{Dimension Reduction for Aerodynamic Design Optimization.}
\newblock {\em AIAA Journal\/}, 49, 6, 1256--1266.

\bibitem[\protect\citename{Vivarelli and Williams, }1999]{Vivarelli:1999}
Vivarelli, F. and Williams, C. K.~I. (1999).
\newblock \enquote{Discovering Hidden Features with {Gaussian} Processes
  Regression.}
\newblock In {\em Proceedings of the 1998 Conference on Advances in Neural
  Information Processing Systems II\/},  613--619. MIT Press.

\bibitem[\protect\citename{Wang et~al., }2016]{Wang2016}
Wang, Z., Hutter, F., Zoghi, M., Matheson, D., and {De Freitas}, N. (2016).
\newblock \enquote{Bayesian Optimization in a Billion Dimensions via Random
  Embeddings.}
\newblock {\em Journal of Artificial Intelligence Research\/}, 55, 1, 361--387.

\bibitem[\protect\citename{{Wolfram Research, Inc}, }2019]{Mathematica}
{Wolfram Research, Inc} (2019).
\newblock \enquote{Mathematica, {V}ersion 12.0.}
\newblock Champaign, IL, 2019.

\end{thebibliography}
	\bibliographystyle{jasa}

	\bigskip
	\begin{center}
		{\large\bf SUPPLEMENTARY MATERIAL}
	\end{center}
	
	\begin{description}
		
		\item[Additional Kernel Expressions and Derivation:] Detailed update
		derivations and kernel expressions for Mat\'ern 3/2 and 5/2 kernels as
		well as gradients for all kernel expressions. (PDF)
		
		\item[R-package for  Sequential Active Subspace UQ:]
		\texttt{R}-package \texttt{activegp} containing code implementing
		methods described in this article (also available from CRAN). (GNU Tar file).
		
	\end{description}

	\appendix
	
	\section{Kernel Expressions}
	
	\label{ap:kernelexps}
	Here we provide specific formulae to evaluate $\int_\Xset \frac{\partial
	k(\x, \x_1)}{\partial x_i} \cdot \frac{\partial k(\x_2, \x)}{\partial x_j}
	d\mu(\x)$, that is, the element $[\Wij]_{1,2}$ of the matrix $\Wij$ used
	in Theorem~\ref{th:C_GP}. Derivation of these formulae was aided by
	Wolfram Mathematica \cite{Mathematica}. $[\Wij]_{1, 2}$ may be written as
	a product of marginal functions as such if $i = j$:
	\begin{equation*}
	[\Wij]_{1, 2} = w_{i,i}([\x_1]_i, [\x_2]_i)
	\prod_{l=1, l\neq i}^{\nvar} I_{l,l}([\x_1]_{l}, [\x_2]_l)
	\end{equation*}
	and, if $i \neq j$:
	\begin{equation*}
	[\Wij]_{1, 2} = w_{i,j}([\x_1]_i, [\x_2]_i)w_{i,j}([\x_2]_j, [\x_1]_j) 
	\prod_{l=1, d\neq i,j}^{\nvar} I_{l,l}([\x_1]_{l}, [\x_2]_l)
	\end{equation*}
	with 
	\begin{equation*}
	w_{i,i}(a, b) = \int_0^1 \frac{\partial k(x,a)}{\partial x} \frac{\partial k(b,x)}{\partial x} dx
	\end{equation*}
	\begin{equation*}
	w_{i,j}(a, b) = \int_0^1 \frac{\partial k(x,a)}{\partial x} k(b,x) dx
	\end{equation*}
	\begin{equation*}
	I_{d,d}(a,b) = \int_0^1 k(x,a)k(b,x)dx.
	\end{equation*}


	Gradients of $\Wij$ with respect to designs (here, $\x_1$ or $\x_2$) are
	available by simply replacing the appropriate term in the product by its
	derivative.

    Closed-form expressions and gradients are available for certain $k$.
    Expressions for the Gaussian, Mat\'ern 3/2 and 5/2 kernels, and gradients
    may be found in the Supplementary Materials; and \texttt{C++} code
    implementing all expressions is available in the \texttt{R} package
    \texttt{activegp}.

	\section{Derivatives of the Updates}
	\label{sec:update}

	We want to express $\C^{(n + 1)}$ as a function of $\CGPn$, $\x_{n+1}$ and
	$y_{n+1}$. Ultimately we are also interested in its gradient. Reusing
	notations of Section~\ref{sec:CGP}, $C_{i,j}^{(n+1)}$ is composed of three
	parts, $E_{i,j}$, $tr(\K_{n+1}^{-1}\Wijnew)$, and $\y_{n+1}^\top
	\K_{n+1}^{-1} \Wijnew \K_{n+1}^{-1} \y^\top_{n+1}$.

	The first component is unchanged with the update. The second one can be
	derived by adapting results from \cite{Binois2018} since $\Wijnew$ may not
	be symmetric: $$tr(\K_{n+1}^{-1}\Wijnew) = tr(\K_n^{-1} \Wijn)
	- \sigma_n^2(\xnew) \vecg(\xnew)^\top \Wijn \vecg(\xnew) - (\vecw_a(\xnew) + \vecw_b(\xnew)) ^\top \vecg(\xnew)
	- \sigma_n^2(\xnew)^{-1} w(\xnew, \xnew).$$
	
	For the remaining one,  partition inverse equations \citep{barnett:1979}
	give
	\begin{equation*}
	\K_{n+1}^{-1}=\begin{bmatrix}
	\K_{n}^{-1} + \vecg(\xnew) \vecg(\xnew)^ \top \sigma_n^2(\xnew) & \vecg(\xnew) \\
	\vecg(\xnew)^\top & \sigma_n^2(\xnew)^{-1}
	\end{bmatrix},
	\label{eq:Knp1i_up}
	\end{equation*}
	where $\vecg(\xnew)= -\sigma_n^2(\xnew)^{-1}\K_n^{-1} \veckn(\xnew)$ and
	$\sigma^2_n(\xnew) = k_n(\xnew, \xnew)$ as in \eqref{eq:kn}. Denote $\G =
	\vecg(\xnew) \vecg(\xnew)^ \top \sigma_n^2(\xnew)$.

	Decomposing and regrouping terms, we get the following.
	\[
	\begin{bmatrix}
	\y_n^\top & y_{n+1}
	\end{bmatrix}
	\begin{bmatrix}
	\K_{n}^{-1}+ \G & \vecg(\xnew) \\
	\vecg(\xnew)^\top & \sigma_n^2(\xnew)^{-1}
	\end{bmatrix}
	\begin{bmatrix}
	\Wijn & \vecw_a(\xnew)\\
	\vecw_b(\xnew)^\top & w(\xnew, \xnew)
	\end{bmatrix}
	\begin{bmatrix}
	\K_{n}^{-1}+ \G & \vecg(\xnew) \\
	\vecg(\xnew)^\top & \sigma_n^2(\xnew)^{-1}
	\end{bmatrix}
	\begin{bmatrix}
	\y_n \\ y_{n+1}
	\end{bmatrix}
	\]

	\[
	= 
	\begin{bmatrix}
	\y_n^\top (\K_{n}^{-1}+ \G) + y_{n+1} \vecg(\xnew)^\top & \y_n^\top \vecg(\xnew) + y_{n+1} \sigma_n^2(\xnew)^{-1}
	\end{bmatrix}
	\Wijnew
	\begin{bmatrix}
	(\K_{n}^{-1}+ \G) \y_n + y_{n+1} \vecg(\xnew)\\
	\vecg(\xnew)^\top \y_n + y_{n+1} \sigma_n^2(\xnew)^{-1}
	\end{bmatrix}
	\]
	\begin{align*}
	=& \y_n^\top (\K_{n}^{-1}+ \G) \Wijn (\K_{n}^{-1}+ \G) \y_n + 
	y_{n+1} \y_n^\top (\K_{n}^{-1}+ \G) \Wijn \vecg(\xnew) + y_{n+1} \vecg(\xnew)^\top \Wijn (\K_{n}^{-1}+ \G) \y_n\\
	&+ y_{n+1}^2 \vecg(\xnew)^\top \Wijn \vecg(\xnew) 
	+ \y_n^\top \vecg(\xnew) \vecw_b(\xnew)^\top (\K_{n}^{-1}+ \G) \y_n +  \y_n^\top \vecg(\xnew) \vecw_b(\xnew)^\top \vecg(\xnew) y_{n+1}\\
	& + y_{n+1} \sigma_n^2(\xnew)^{-1} \vecw_b(\xnew)^\top (\K_{n}^{-1}+ \G) \y_n + y_{n+1}^2 \sigma_n^2(\xnew)^{-1} \vecw_b(\xnew)^\top \vecg(\xnew)\\
	&+ \y_n^\top (\K_{n}^{-1}+ \G)\vecw_a(\xnew)\vecg(\xnew)^\top \y_n + y_{n+1} \sigma_n^2(\xnew)^{-1} \y_n^\top (\K_{n}^{-1}+ \G) \vecw_a(\xnew) + y_{n+1} \vecg(\xnew)^\top \vecw_a(\xnew) \vecg(\xnew)^\top \y_n \\ &+ y_{n+1}^2 \sigma_n^2(\xnew)^{-1} \vecg(\xnew)^\top \vecw_a(\xnew)
	+  w(\xnew, \xnew) \y_n^\top \vecg(\xnew) \vecg(\xnew)^\top \y_n +  w(\xnew, \xnew) y_{n+1} \sigma_n^2(\xnew)^{-1} \y_n^\top \vecg(\xnew)\\ &+  y_{n+1} \sigma_n^2(\xnew)^{-1} w(\xnew, \xnew) \vecg(\xnew)^\top \y_n + y_{n+1}^2 \sigma_n^2(\xnew)^{-2} w(\xnew, \xnew)
	\end{align*}

	\resp{SW}{After replacing $\G$ and $\vecg$ by their expressions and factoring}, 
	we get the expression for $C_{i,j}^{(n+1)} - C_{i,j}^{(n)}=$
	\begin{align*}
	& - (\vecw_a(\xnew) + \vecw_b(\xnew))^\top \vecg(\xnew)\\ 
	&- (\y_n^\top \vecg(\xnew) + y_{n+1} \sigma_n^2(\xnew)^{-1}) \left[ \y_n^\top \K_n^{-1} \Wijn \K_n^{-1} \veckn(\xnew) + \veckn(\xnew)^\top \K_n^{-1} \Wijn \K_n^{-1} \y_n \right]\\ 
	&+ (\y_n^\top \vecg(\xnew) + y_{n+1} \sigma_n^2(\xnew)^{-1}) (\vecw_a(\xnew) + \vecw_b(\xnew))^\top (\K_{n}^{-1} \y_n + \vecg(\xnew) y_{n+1} - \vecg(\xnew) \veckn(\xnew)^\top \K_n^{-1} \y_n)\\
	&+ \left[ (\y_n^\top \vecg(\xnew) + y_{n+1} \sigma_n^2(\xnew)^{-1})^2 - \sigma_n^2(\xnew)^{-1} \right] \left[ w(\xnew, \xnew) + \veckn(\xnew)^\top \K_n^{-1} \Wijn \K_n^{-1} \veckn(\xnew) \right].
	\end{align*}

	\section{Infill Criteria Derivation}
	\label{sec:infill}

	At step $n$, $y_{n+1} \sim \mathcal{N}(m_n(\xnew), \sigma_n^2(\xnew))$.
	Taking the expectation with respect to $y_{n+1}$ considerably simplify the
	expressions: $$\Esp{\omega}{C_{i,j}^{(n+1)} | \A_n } = C_{i,j}^{(n)}$$
	$$\Var{\omega}{ C_{i,j}^{(n+1)} | \A_n } = \beta^2 + 2 \gamma^2,$$
	\resp{SW}{where expressions for $\beta$ and $\gamma$ are given below. 
	Indeed:}\\ 1) $\Esp{\omega}{
	\y_n^\top \vecg(\xnew) + y_{n+1} \sigma_n^2(\xnew)^{-1} } =
	(\y_n^\top \vecg(\xnew) + m_n(\xnew) \sigma_n^2(\xnew)^{-1})
	= 0$ by definition of $\vecg(\xnew)$\\
	2) $\Esp{\omega}{ (\y_n^\top \vecg(\xnew) + y_{n+1} \sigma_n^2(\xnew)^{-1})^2} 
	= \sigma_n^2(\xnew)^{-2} m_n(\xnew) -2 \sigma_n^2(\xnew)^{-2} m_n(\xnew)^2 +
	 \sigma_n^2(\xnew)^{-2} (m_n(\xnew)^2 + \sigma_n^2(\xnew)) 
	= \sigma_n^2(\xnew)^{-1}$\\
	3) $\Esp{\omega}{ y_{n+1} (\y_n^\top \vecg(\xnew) + y_{n+1} \sigma_n^2(\xnew)^{-1})}
	= -m_n(\xnew)^2 \sigma_n^2(\xnew)^{-1} + \sigma_n^2(\xnew)^{-1}(m_n(\xnew)^2 +
	 \sigma_n^2(\xnew)) = 1$.

	Notice that if  we use the plugin estimator of $y_{n+1}$, namely,
	$m_n(\xnew)$, then $$\hat{C}_{i,j}^{(n+1)} = C_{i,j}^{(n)} -
	\left(\sigma_n^2(\xnew)^{-1} \veckn(\xnew)^\top \K_n^{-1} \Wijn \K_n^{-1}
	\veckn(\xnew) + (\vecw_a(\xnew) + \vecw_b(\xnew))^\top \vecg(\xnew) +
	\sigma_n^2(\xnew)^{-1} w(\xnew, \xnew) \right),$$ which reduces to the
	integrated mean square prediction error of the corresponding derivatives.

	For further simplifications, denote by $Z = \frac{y_{n+1} -
	m_n(\xnew)}{\sigma_n(\xnew)}$ \resp{SW}{a} $\mathcal{N}(0, 1)$
	random variable.

	Then $C_{i,j}^{(n+1)} - C_{i,j}^{(n)}$ can be rewritten as
	\begin{align*}
	& - (\vecw_a(\xnew) + \vecw_b(\xnew))^\top \vecg(\xnew) - \sigma^2_n(\xnew)^{-1} \left[ w(\xnew, \xnew) + \veckn(\xnew)^\top \K_n^{-1} \Wijn \K_n^{-1} \veckn(\xnew) \right]\\ 
	&+ Z \sigma_n(\xnew)^{-1} \left[ \y_n^\top \K_n^{-1} \Wijn \K_n^{-1} \veckn(\xnew) + \veckn(\xnew)^\top \K_n^{-1} \Wijn \K_n^{-1} \y_n - (\vecw_a(\xnew) + \vecw_b(\xnew))^\top \K_{n}^{-1} \y_n \right]\\ 
	&+  Z^2 \sigma_n^2(\xnew)^{-1} \left[ w(\xnew, \xnew) + \veckn(\xnew)^\top 
	\K_n^{-1} \Wijn \K_n^{-1} \veckn(\xnew)- (\vecw_a(\xnew) + \vecw_b(\xnew))^\top 
	\K_n^{-1} \veckn(\xnew) \right].
	\end{align*}

	Now denote the coefficient to the linear $Z$ term as $\beta_{i,j}$ and
	that to the quadratic $Z$ term as $\gamma_{i,j}$, and further denote the
	matrices containing these as entries given $\xnew$ as $\mathbf{B}(\xnew)$
	and $\bm{\Gamma}(\xnew)$, respectively.  We further define 3D arrays
	$\partial \mathbf{B}(\xnew)$ and $\partial \bm{\Gamma}(\xnew)$ such that
	$\partial \mathbf{B}(\xnew)_d$ gives a matrix of derivatives of each
	element of $\mathbf{B}(\xnew)$ with respect to the $d$th element of
	$\xnew$:
	\begin{equation*}
	\partial \mathbf{B}(\xnew)_d =\y_n^\top\K_n^{-1}(\Wijn + (\Wijn)^\top) 
	\K_n^{-1} \boldsymbol{\kappa}_d(\xnew)^\top - 
	\y_n^\top\K_n^{-1}\left(\frac{\partial \vecw_a(\xnew)}{\partial \xnew_d} + 
	\frac{\partial \vecw_b(\xnew)}{\partial \xnew_d}\right) - 
	\mathbf{B} \frac{\partial\sigma_n(\xnew)}{\partial \xnew_d} \sigma_n(\xnew)^{-\frac{1}{2}}
	\end{equation*}
	and
	\begin{multline*}
	\partial \bm{\Gamma}(\xnew)_d = \sigma_n^{-1}(\xnew)\Big(\frac{\partial 
	w(\xnew, \xnew)}{\xnew_d}^\top + 
	\veckn(\xnew)^\top \K_{n}^{-1}(\Wijn + (\Wijn)^\top) \K_{n}^{-1} 
	\boldsymbol{\kappa}_d(\xnew)^\top - \\
	\veckn(\xnew)^\top\K_{n}^{-1}\left(\frac{\partial \vecw_a(\xnew)}{\partial 
	\xnew_d} + \frac{\partial \vecw_b(\xnew)}{\partial \xnew_d}\right) - 
	(\vecw_a(\xnew) + \vecw_b(\xnew))^\top \K_n^{-1} \frac{\partial w(\xnew, \xnew)}{\xnew_d}
	- \frac{\partial\sigma_n(\xnew)}{\partial \xnew_d} \bm{\Gamma}
	\Big).
	\end{multline*}

	We are now equipped to derive closed-form expressions for our infill
	criteria, as well as gradients of those expressions with respect to design
	locations, the results of which are given in Table~\ref{tab:acq_expr}.


\vspace{3em}

\small

\framebox{\parbox{\linewidth}{
The submitted manuscript has been created by UChicago Argonne, LLC, Operator of 
Argonne National Laboratory (``Argonne''). Argonne, a U.S.\ Department of 
Energy Office of Science laboratory, is operated under Contract No.\ 
DE-AC02-06CH11357. 
The U.S.\ Government retains for itself, and others acting on its behalf, a 
paid-up nonexclusive, irrevocable worldwide license in said article to 
reproduce, prepare derivative works, distribute copies to the public, and 
perform publicly and display publicly, by or on behalf of the Government.  The 
Department of Energy will provide public access to these results of federally 
sponsored research in accordance with the DOE Public Access Plan. 
http://energy.gov/downloads/doe-public-access-plan.}}

\end{document}